%% file: preprint.tex
\documentclass{article}
\usepackage[preprint]{colm2025_conference}

\usepackage{microtype}
\usepackage[utf8]{inputenc} 
\usepackage[T1]{fontenc}    
\definecolor{green}{RGB}{0,150,10}
\definecolor{blue}{RGB}{0,148,181}
\definecolor{orange}{RGB}{194,153,107}
\usepackage[colorlinks,linkcolor=red,anchorcolor=green,citecolor=blue]{hyperref}
\usepackage{url}            

\usepackage{lineno}

\usepackage{booktabs}       
\usepackage{amsfonts}       
\usepackage{nicefrac}       
\usepackage{xcolor}         
\usepackage{amsmath}
\usepackage{marvosym}
\usepackage{graphicx}
\usepackage{colortbl}
\usepackage{multirow}
\usepackage{subcaption}
\usepackage{changes}
\usepackage{listings}
\usepackage{mdframed}
\usepackage{caption}
\usepackage{booktabs}
\usepackage{colortbl}
\usepackage{multicol}

\usepackage[capitalize,noabbrev]{cleveref}

\title{InternAgent: When Agent Becomes the Scientist – Building Closed-Loop System from Hypothesis to Verification}

\newcommand{\homepage}{\raisebox{-1.5pt}{\includegraphics[height=1em]{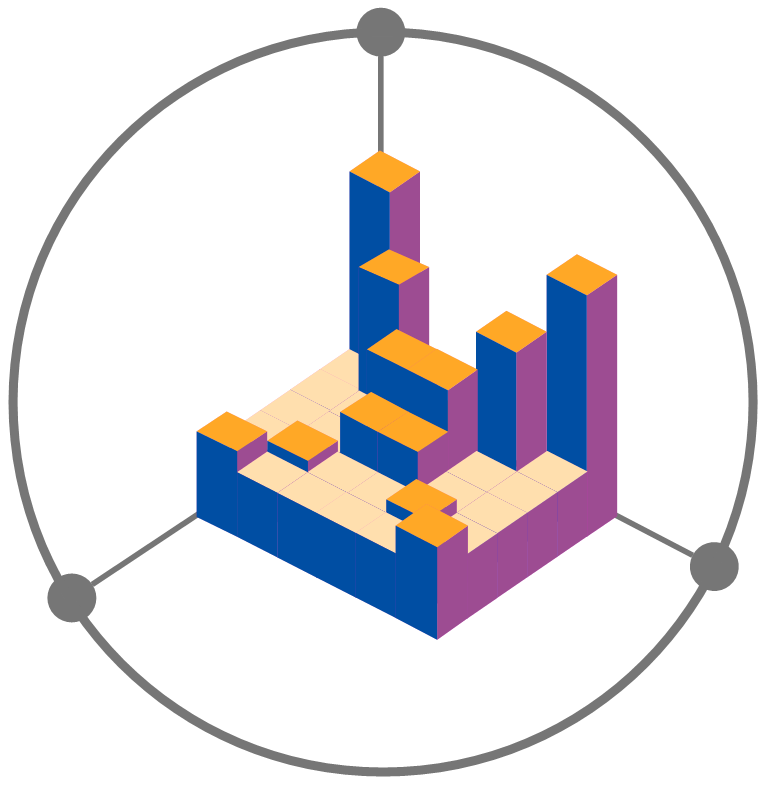}}}
\newcommand{\github}{\raisebox{-1.5pt}{\includegraphics[height=1em]{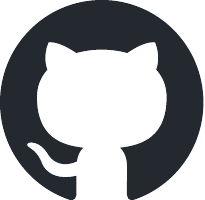}}}
\newcommand{\huggingface}{\raisebox{-1.5pt}{\includegraphics[height=1em]{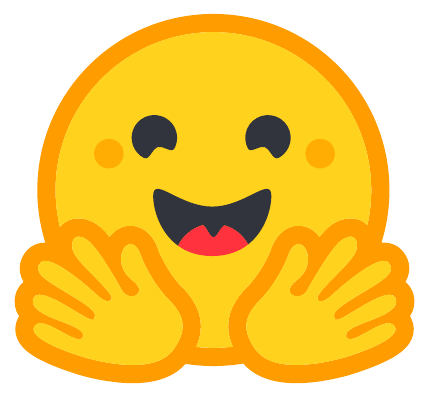}}}

\author{%
InternAgent Team, Shanghai Artificial Intelligence Laboratory \\ [1.5mm]
{\homepage\ \texttt{\url{https://alpha-innovator.github.io/InternAgent-project-page/}}} \\
{\github\ \texttt{\url{https://github.com/Alpha-Innovator/InternAgent}}} \\
{\huggingface\ \texttt{\url{https://huggingface.co/U4R/InternAgent}}}
}

\begin{document}

\maketitle

\begin{figure}[h]
\vspace{-16pt}
    \centering
    \includegraphics[width=\linewidth]{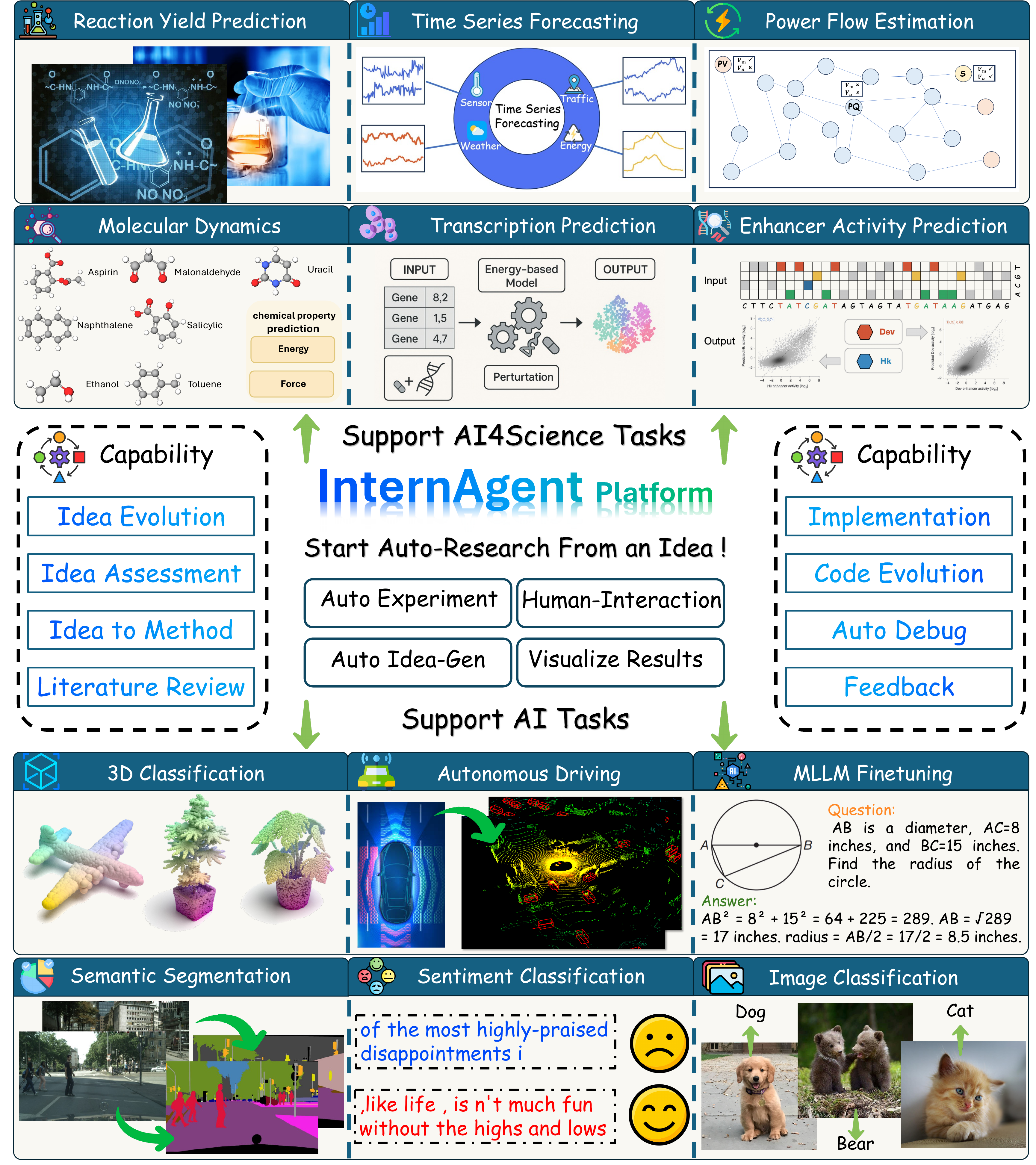}
    \vspace{-10pt}\caption{\textbf{InternAgent} can support \textbf{12} types of scientific research tasks ranging from the AI field to the science field, including reaction yield prediction, molecular dynamics, power flow estimation, time series forecasting, transcription prediction, enhancer activity prediction, sentiment classification, 2D image classification, 3D point classification, 2D semantic segmentation, 3D autonomous driving, large vision-language model fine-tuning.}
    \label{fig:figure1}
\end{figure}

\input{sec/0_abstract}
\input{sec/1_intro}

\input{sec/2_framework}
\input{sec/3_quantitative}
\input{sec/5_qualitative}
\input{sec/6_related_works}

\section{Conclusion and Future Works}
\noindent \textbf{Summary.} We have introduced a closed-loop multi-agent framework for the first time, which supports 12 types of scientific research tasks. It has been validated to generate novel ideas and transform these ideas into code that can effectively improve performance. InternAgent refines the initially generated ideas through human-interactive feedback enriched with a self-evolutionary path of ideas. It facilitates the transformation from coarse-grained proposals to fine-grained methodologies via an idea-to-methodology construction process. Furthermore, by leveraging multi-round experimental planning and execution, it implements the corresponding theoretical methods, thereby completing the closed-loop process in scientific research—from hypothesis generation to verification.

\noindent \textbf{Future Outlook.} InternAgent faces several key technical challenges that need to be addressed in the future:
\begin{itemize}
    \item Knowledge Retrieval: This involves establishing connections and relationships between papers, utilizing deep research techniques to conduct extensive searches across a wide range of papers, and performing meta-analyses on the search results. Additionally, it requires transforming the papers into structured representations such as triples, and utilizing graph networks to uncover relationships between papers, including associations in paper ideas, methodologies, experimental conditions, and experimental results. Moreover, retrieval-augmented generation will be employed to alleviate the hallucination issues of LLMs when generating viewpoints or citing references.
    \item Knowledge Understanding and Representation: This involves utilizing VLM/LLM to accurately analyze relevant academic papers, aiming to understand the significance of their core concepts, methodologies, and research findings, while also refining knowledge and formulating hypotheses. Additionally, it focuses on extracting valuable knowledge from a large number of papers, identifying common patterns, trends, and connections, thereby advancing the understanding and representation of knowledge in the field.
    \item Agent Capability Enhancement: This focuses on improving the ability of AI systems to autonomously perform complex tasks in scientific research. The strength of agents lies in their ability to dynamically adapt, rather than solely relying on historical records to determine subsequent actions. Through self-modification, they can flexibly redefine their initial goals and planning strategies while utilizing feedback, as well as communication logs between agents or between humans and agents, to train and improve themselves. This mechanism should focus on improving their ability to gather feedback from three key sources: the environment, interactions with other agents, and human experts.
    \item Scientific Discovery-related Benchmark Construction: This involves evaluating the value that an idea can bring, rather than simply evaluating its novelty. It also includes evaluating whether the methods proposed by AI align with their corresponding code implementations and determining whether InternAgent demonstrates a certain level of generalization ability in broader scientific scenarios.

\end{itemize}

\bibliography{main}
\bibliographystyle{colm2025_conference}

\newpage
\appendix

\section*{Appendix}

\vspace{6pt}
\section{Contributions and Acknowledgments}

\definecolor{damaiblue}{RGB}{10, 102, 155}
\definecolor{damaiorange}{RGB}{180,50,50}
\definecolor{damaired}{RGB}{10, 50, 50}

\vspace{4pt}
\begin{multicols}{2} %
\noindent
\textbf{\color{damaired} 1. Core Contributors} \\ [1.5mm]
\color{damaired} Bo Zhang \\  
\color{damaired} Shiyang Feng \\
\color{damaired} Xiangchao Yan \\
\color{damaired} Jiakang Yuan \\

\noindent
\textbf{\color{damaired} 2. General Multi-agent System} \\ [1.5mm]
\color{damaired} \textbf{Deep Research Agent}: Yusong Hu, Runmin Ma, Bo Zhang \\
\color{damaired} \textbf{Self-Evolving Idea Agent}: Shiyang Feng \\
\color{damaired} \textbf{Idea-to-Methodology Agent}: Xiangchao Yan \\
\color{damaired} \textbf{Coding Agent}: Shiyang Feng, Jiakang Yuan, Xiangchao Yan \\
\color{damaired} \textbf{Multi-round Experimental Plan}: Jiakang Yuan, Bo Zhang, Xiangchao Yan, Shiyang Feng \\

\noindent
\textbf{\color{damaired} 3. Scientific Research Task Integration and Enhancement} \\  [1.5mm]
\color{damaired} \textbf{Reaction Yield Prediction}: Xiaohan He, Zhiyin Yu\\
\color{damaired} \textbf{Molecular Dynamics}: Zhiyin Yu\\
\color{damaired} \textbf{Power
Flow Estimation}: Songtao Huang\\
\color{damaired} \textbf{Time Series Forecasting}: Songtao Huang\\
\color{damaired} \textbf{Sentiment CLS}: Xiangchao Yan, Jiakang Yuan\\
\color{damaired} \textbf{3D Point CLS}: Xiangchao Yan, Jiakang Yuan\\
\color{damaired} \textbf{2D Image CLS}: Xiangchao Yan, Jiakang Yuan\\
\color{damaired} \textbf{2D Semantic Segmentation}: Shiyang Feng \\
\color{damaired} \textbf{3D Autonomous Driving}: Jiakang Yuan \\
\color{damaired} \textbf{Transcription Prediction}: Xiaohan He, Peng Ye \\
\color{damaired} \textbf{Enhancer Activity Prediction}: Zhiyin Yu \\
\color{damaired} \textbf{VLM finetuning}: Tianshuo Peng \\

\noindent
\textbf{\color{damaired} 4. Software Development} \\  [1.5mm]
\color{damaired} Zheng Nie, Zhilong Wang,  Runmin Ma, Jinyao Liu, Shiyang Feng, Xiangchao Yan \\

\noindent
\textbf{\color{damaired} 5. Manuscript Preparation} \\  [1.5mm]
\color{damaired} Shiyang Feng, Bo Zhang, Xiangchao Yan, Jiakang Yuan, Zhiyin Yu, Songtao Huang, Lei Bai, Xiaohan He, Tianshuo Peng \\

\noindent
\textbf{\color{damaired} 6. Idea Human Evaluation Specialist} \\  [1.5mm]
\color{damaired} Bo Zhang, Peng Ye, Shufei Zhang, Dongzhan Zhou, Xiaosong Wang, Lei Bai \\

\noindent
\textbf{\color{damaired} 7. Project Management and Product} \\  [1.5mm]
\color{damaired} Yilan Zhang, Meng Li, Shaowei Hou, Zhongying Tu \\

\noindent
\textbf{\color{damaiblue} 8. Advising} \\  [1.5mm]
\color{damaiblue} Bowen Zhou, Wanli Ouyang, Xiangyu Yue \\

\noindent
\textbf{\color{damaiblue} 9. Project Co-lead} \\  [1.5mm]
\color{damaiblue} Lei Bai, bailei@pjlab.org.cn \\
\color{damaiblue} Bo Zhang, zhangbo@pjlab.org.cn \\

\end{multicols} %

\section{Evaluation Details}

\subsection{Scoring Criteria for Idea Review}
\label{app:criteria}
In Table~\ref{tab:idea_human_eval} of the main text, we conducted a human evaluation to assess the novelty of ideas generated by AI-Scientist-V2~\citep{yamada2025ai} and our InternAgent. The evaluation was carried out across four dimensions: soundness, contribution, overall rating, and confidence.

Specifically, considering the evaluation cost, we opted to evaluate four types of research tasks: reaction yield prediction, 2D semantic segmentation, 2D image classification, and point cloud autonomous driving. Each invited researcher was required to have peer-review qualifications for top-tier journals or conferences in the relevant field. For each research task, we generated 20 ideas using both AI-Scientist-V2 and InternAgent, and five experienced researchers were invited to score each idea. In this part, we provide a detailed description of the scoring criteria for each reviewer, as outlined below:

\textbf{Soundness}:
\begin{itemize}
  \item 4 excellent
  \item  3 good
  \item  2 fair
  \item  1 poor
\end{itemize}

\textbf{Contribution}:
\begin{itemize}
  \item 4 excellent
  \item  3 good
  \item  2 fair
  \item  1 poor
\end{itemize}

\textbf{Overall}:
\begin{itemize}
  \item Rating: 10: Award quality: Technically flawless paper with groundbreaking impact on one or more areas of AI, with exceptionally strong evaluation, reproducibility, and resources, and no unaddressed ethical considerations.
  \item Rating: 9: Very Strong Accept: Technically flawless paper with groundbreaking impact on at least one area of AI and excellent impact on multiple areas of AI, with flawless evaluation, resources, and reproducibility, and no unaddressed ethical considerations.
  \item Rating: 8: Strong Accept: Technically strong paper with, with novel ideas, excellent impact on at least one area of AI or high-to-excellent impact on multiple areas of AI, with excellent evaluation, resources, and reproducibility, and no unaddressed ethical considerations.
  \item Rating: 7: Accept: Technically solid paper, with high impact on at least one sub-area of AI or moderate-to-high impact on more than one area of AI, with good-to-excellent evaluation, resources, reproducibility, and no unaddressed ethical considerations.
  \item Rating: 6: Weak Accept: Technically solid, moderate-to-high impact paper, with no major concerns with respect to evaluation, resources, reproducibility, ethical considerations.
  \item Rating: 5: Borderline accept: Technically solid paper where reasons to accept outweigh reasons to reject, e.g., limited evaluation. Please use sparingly.
  \item Rating: 4: Borderline reject: Technically solid paper where reasons to reject, e.g., limited evaluation, outweigh reasons to accept, e.g., good evaluation. Please use sparingly.
  \item Rating: 3: Reject: For instance, a paper with technical flaws, weak evaluation, inadequate reproducibility and incompletely addressed ethical considerations.
  \item Rating: 2: Strong Reject: For instance, a paper with major technical flaws, and/or poor evaluation, limited impact, poor reproducibility and mostly unaddressed ethical considerations.
  \item Rating: 1: Very Strong Reject: For instance, a paper with trivial results or unaddressed ethical considerations
\end{itemize}

\textbf{Confidence}:
\begin{itemize}
  \item Confidence: 5: You are absolutely certain about your assessment. You are very familiar with the related work and checked the math/other details carefully.
  \item Confidence: 4: You are confident in your assessment, but not absolutely certain. It is unlikely, but not impossible, that you did not understand some parts of the submission or that you are unfamiliar with some pieces of related work.
  \item Confidence: 3: You are fairly confident in your assessment. It is possible that you did not understand some parts of the submission or that you are unfamiliar with some pieces of related work. Math/other details were not carefully checked.
  \item Confidence: 2: You are willing to defend your assessment, but it is quite likely that you did not understand the central parts of the submission or that you are unfamiliar with some pieces of related work. Math/other details were not carefully checked.
  \item Confidence: 1: Your assessment is an educated guess. The submission is not in your area or the submission was difficult to understand. Math/other details were not carefully checked.
\end{itemize}

\subsection{Information on the Expert Review Process}

\textbf{Qualifications for Human Evaluators}: Evaluators must hold a Ph.D. or be Ph.D. candidates with reviewing experience in top-tier AI conferences such as ICLR, ICML, NeurIPS, CVPR, ICCV, and ACL.

\textbf{Steps for Expert Evaluation and Validation}:
\begin{itemize}
    \item Before the evaluation begins, evaluators are required to carefully read the scoring guidelines, as outlined in Appendix~\ref{app:criteria}.
    
    \item Each evaluator is assigned 20 ideas generated by InternAgent and 20 ideas generated by AI-Scientist-V2~\citep{yamada2025ai}. For each idea, evaluators must carefully review the generated content and provide final scores across four dimensions: Soundness, Contribution, Overall, and Confidence.

    \item Qualified evaluators are required to spend at least 10 minutes reading each idea. During the evaluation process, they are allowed to conduct relevant literature searches and verify idea redundancy to ensure that the scoring results are objective and representative.

\end{itemize}

\begin{figure}[t]
    \centering
    \includegraphics[width=\linewidth]{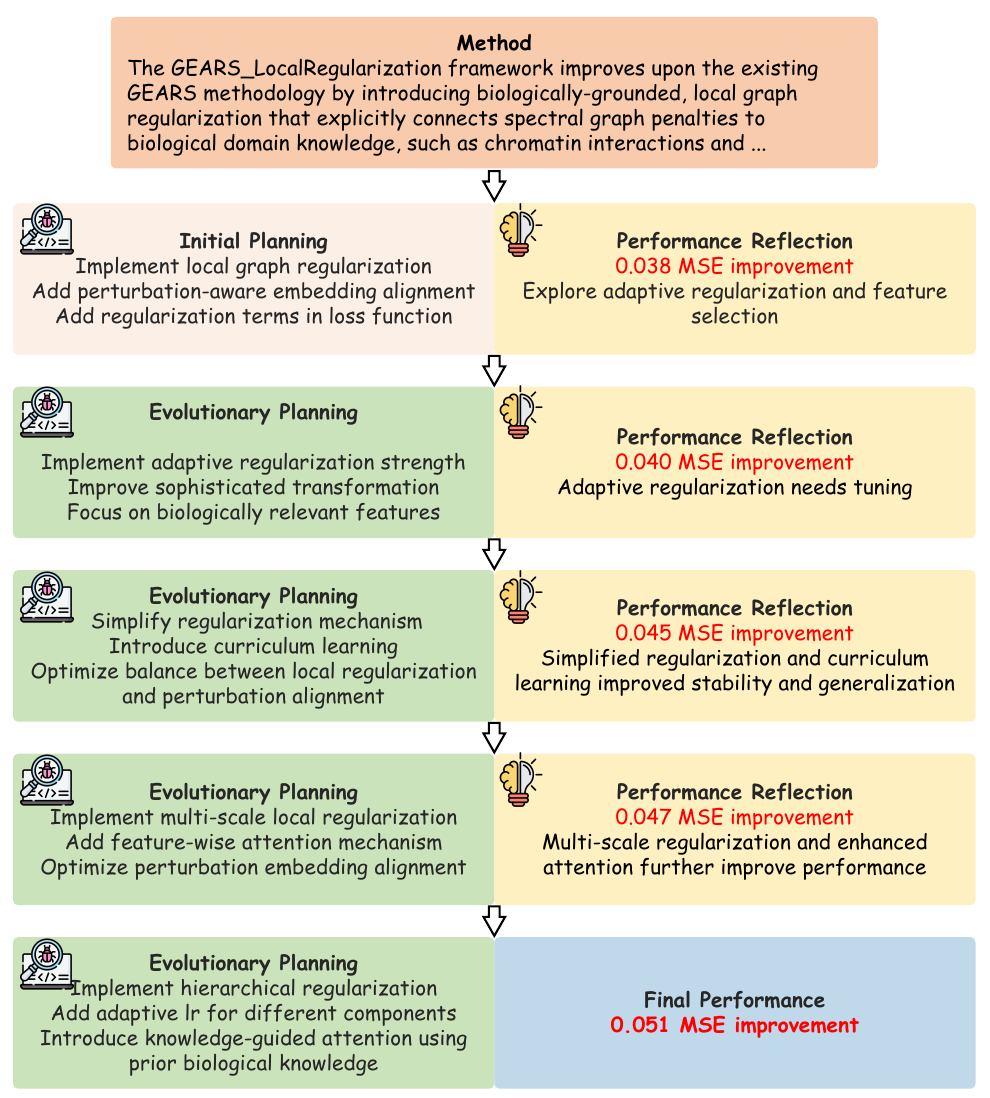}
    \vspace{-10pt}
    \caption{Visual Examples of Experimental Planning and Adaptive Evolution on AutoTPPR task.}
    \label{fig:run5_showcase_tppr}
    \vspace{-10pt}
\end{figure}

\begin{figure}[t]
    \centering
    \includegraphics[width=\linewidth]{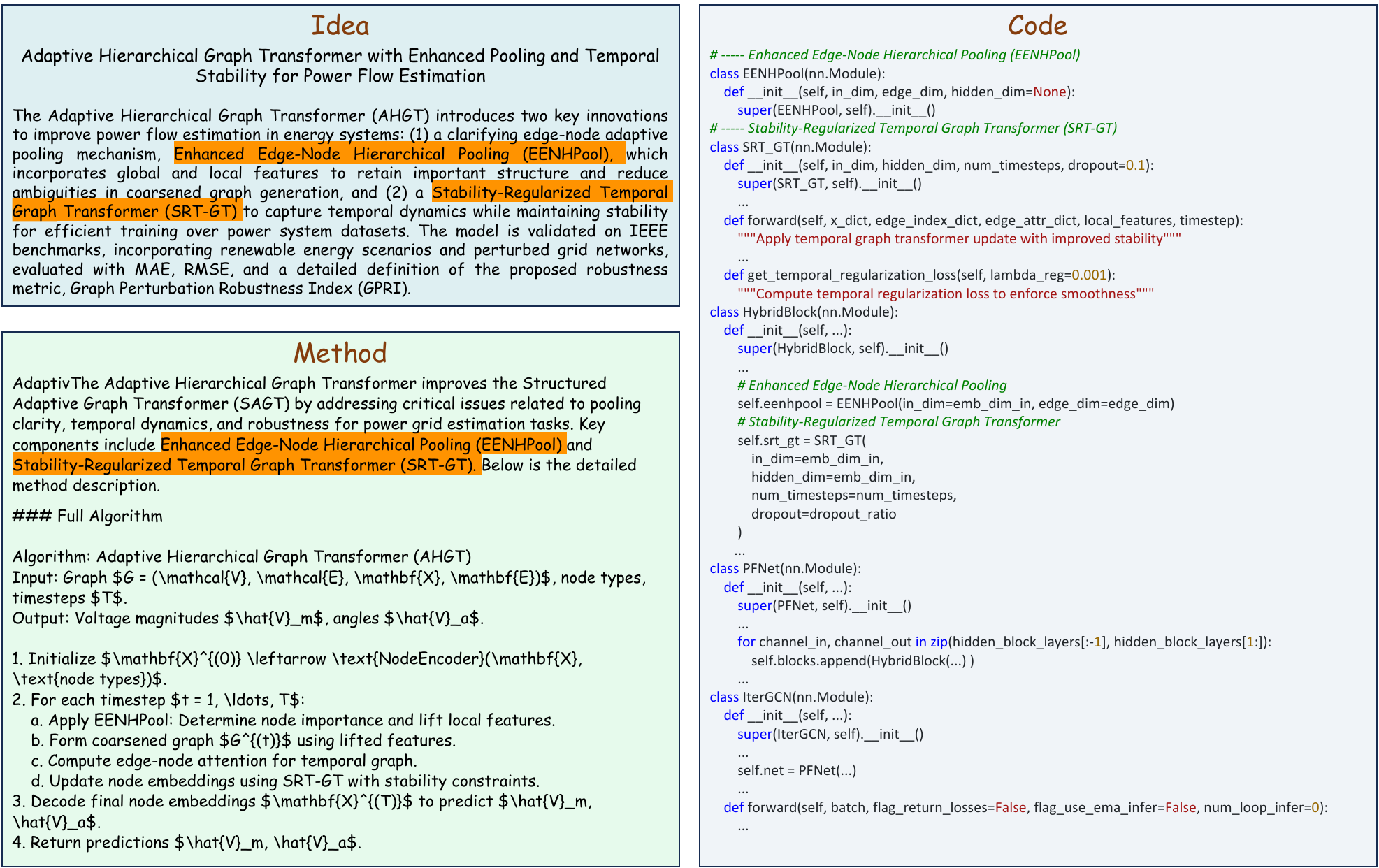}
    \vspace{-10pt}
    \caption{Result Analysis (Generated Idea, Method, and Code) of AutoPower Task}
    \label{fig:showcase_AutoPower}
\end{figure}

\begin{figure}[t]
    \centering
    \includegraphics[width=\linewidth]{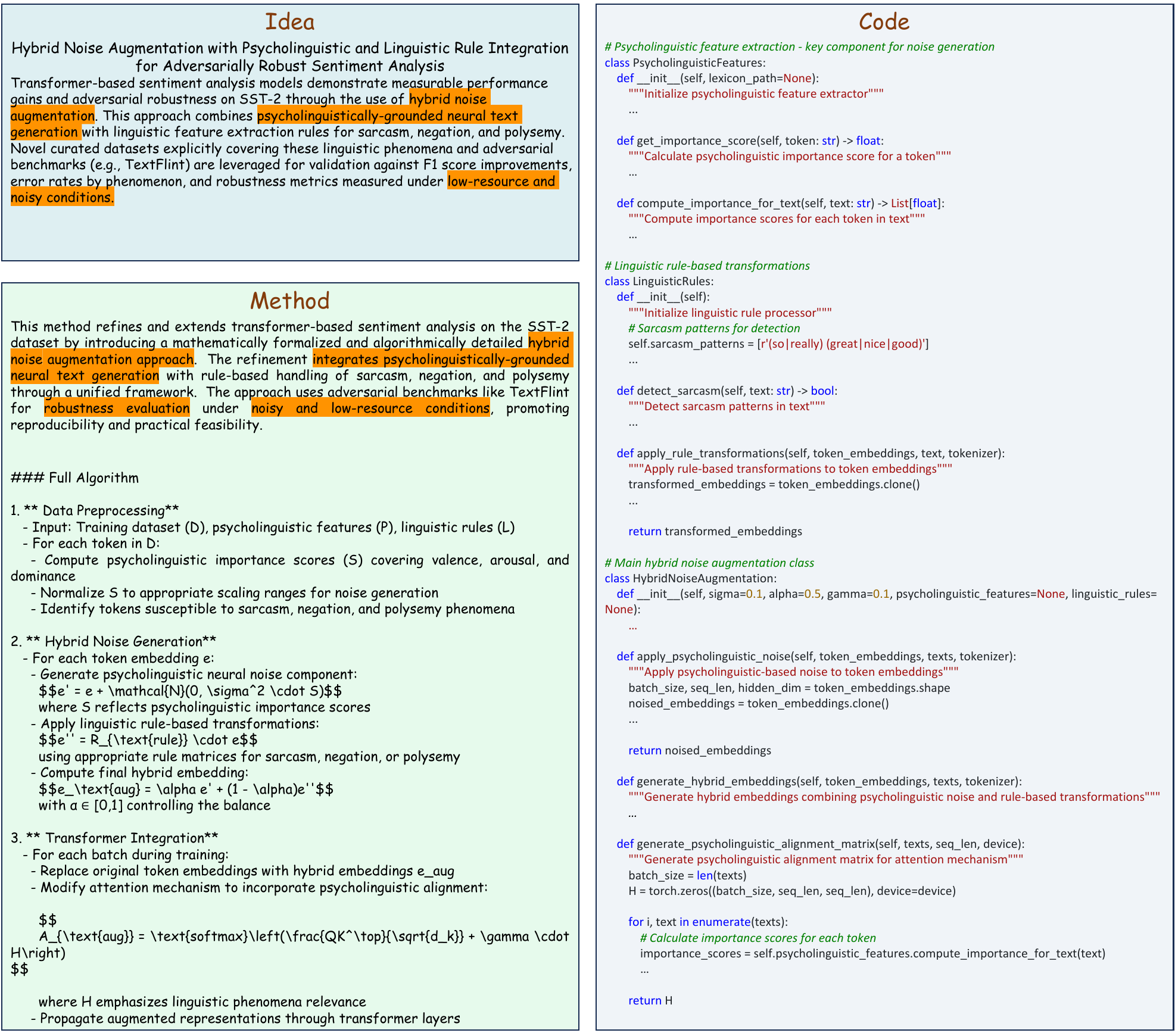}
    \vspace{-18pt}
    \caption{Result Analysis (Generated Idea, Method, and Code) of AutoSenCls Task}
    \label{fig:showcase_AutoSenCls}
\end{figure}

\begin{figure}[t]
    \centering
    \includegraphics[width=\linewidth]{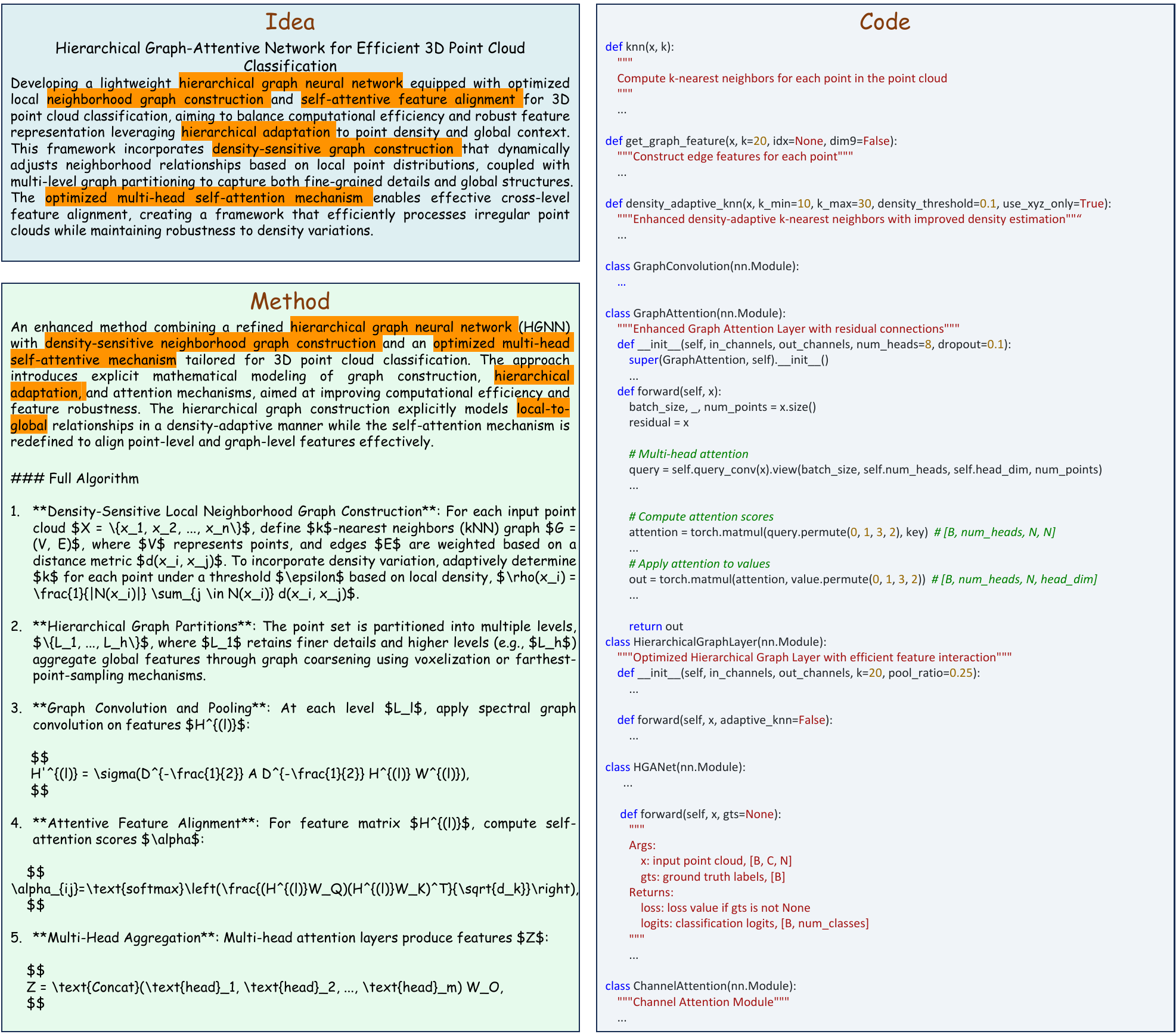}
    \vspace{-18pt}
    \caption{Result Analysis (Generated Idea, Method, and Code) of Auto3DCls Task}
    \label{fig:showcase_Auto2DCls}
\end{figure}

\begin{figure}[t]
    \centering
    \includegraphics[width=\linewidth]{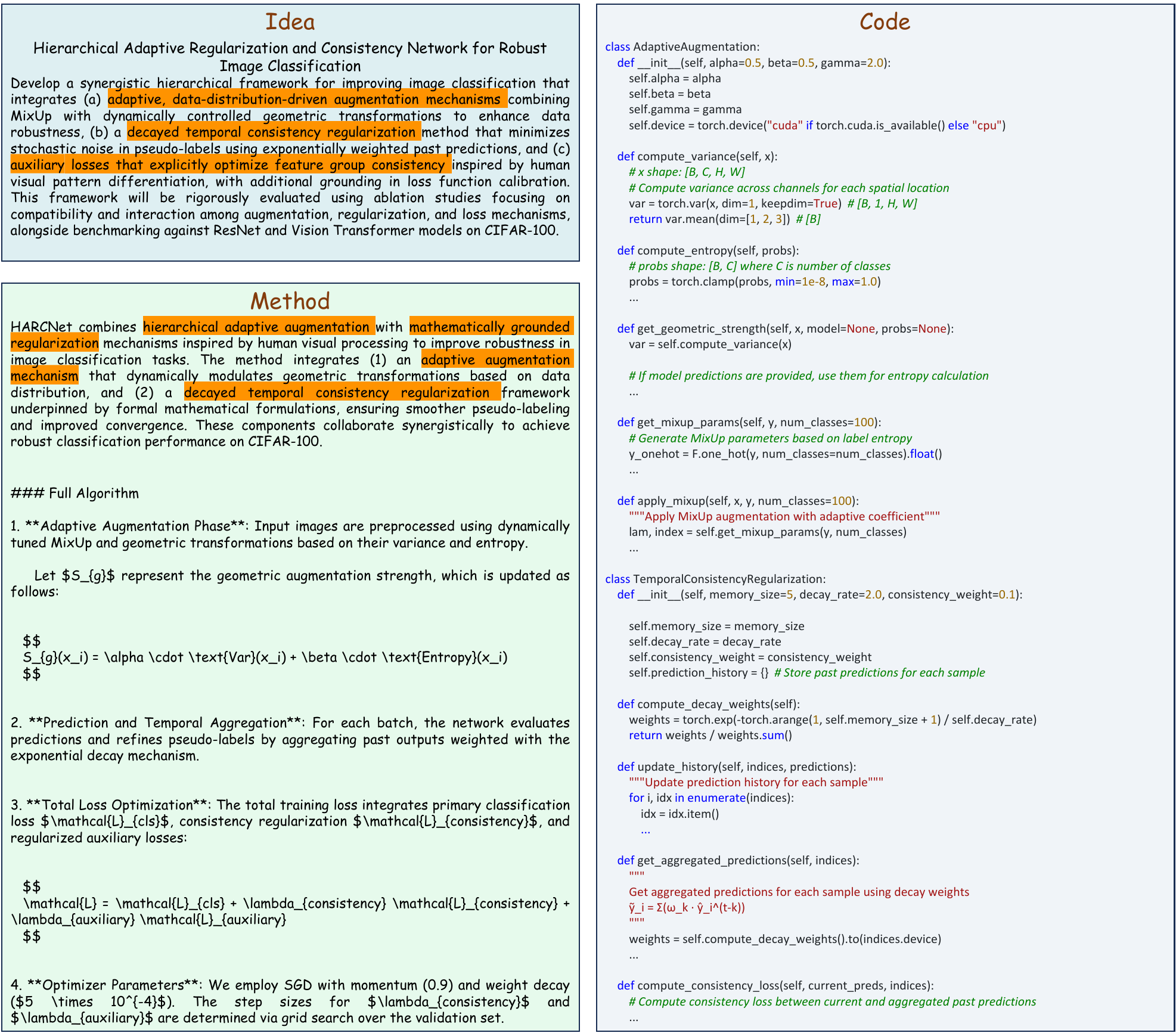}
    \vspace{-18pt}
    \caption{Result Analysis (Generated Idea, Method, and Code) of Auto2DCls Task}
    \label{fig:showcase_Auto3DCls}
\end{figure}

\input{sec/software}

\section{Result Analysis}

We further conducted detailed visualization and analysis around the multi-round experimental planning and execution (Fig.~\ref{fig:run5_showcase_tppr}) in InternAgent, as well as the automated scientific research tasks (Figs.~\ref{fig:showcase_AutoPower},~\ref{fig:showcase_AutoSenCls},~\ref{fig:showcase_Auto2DCls} and~\ref{fig:showcase_Auto3DCls}) supported by InternAgent.

Fig.~\ref{fig:run5_showcase_tppr} illustrates the process of Experimental Planning and Adaptive Evolution on AutoTPPR. Each block in the figure represents a step in the multi-round experiment planning process, where the complete InternAgent-generated method is decomposed into multiple logical steps. This allows for task decomposition during the experimental validation phase, thereby facilitating more significant benchmark results. 

Furthermore, in the AutoPower task illustrated in Fig.~\ref{fig:showcase_AutoPower}, the "Adaptive Hierarchical Graph Transformer" (AHGT) introduces significant advancements for power flow estimation in energy systems. This approach features two key innovations: the Enhanced Edge-Node Hierarchical Pooling (EENHPool) mechanism, which integrates global and local features to retain crucial graph structures while reducing ambiguities, and the Stability-Regularized Temporal Graph Transformer (SRT-GT), designed to capture temporal dynamics while maintaining training stability. These components together enhance the model's robustness and accuracy, validated on IEEE benchmarks under scenarios involving renewable energy and grid perturbations. The AHGT method outputs precise voltage magnitude and angle predictions, assessed using metrics such as MAE, RMSE, and the Graph Perturbation Robustness Index (GPRI). This approach showcases the potential for improved power system modeling through advanced graph transformer techniques.

\end{document}

%% file: sec/0_abstract.tex
\begin{abstract}
Artificial Intelligence (AI) is accelerating the transformation of scientific research paradigms, not only enhancing research efficiency but also driving innovation. We introduce InternAgent, a unified closed-loop multi-agent framework to conduct Autonomous Scientific Research (ASR) across various scientific research fields, enabling researchers to tackle complicated problems in these fields with unprecedented speed and precision. InternAgent highlights three key advantages: 1) \textbf{Scalability}: InternAgent has demonstrated its versatility across \textbf{12} scientific research tasks, capable of generating innovative ideas to enhance the performance of baseline code. 2) \textbf{Interactivity}: InternAgent provides an interface for human expert feedback and multi-agent interaction in automated end-to-end processes, allowing for the seamless integration of domain expert knowledge. 3) \textbf{Efficiency}: InternAgent has achieved promising performance gains in several scientific fields with significantly less time cost compared to human efforts. For instance, in reaction yield prediction, it increased from 27.6\% to 35.4\% in just 12 hours; in enhancer activity prediction, the Pearson correlation coefficient rose from 0.65 to 0.79 with only 4 hours of processing; and in 2D semantic segmentation, precision advanced from 78.8\% to 81.0\% in a mere 30 hours.
\end{abstract}

%% file: sec/1_intro.tex
\section{Introduction}
\label{sec:intro}

Autonomous Scientific Discovery (ASD) refers to the use of Large Language Models (LLMs)~\citep{yang2024qwen2-5,internvl2,qwen2vl,Deepseek-r1} and robotics to independently perform scientific research without direct human intervention~\citep{yuan2025dolphin,yan2025surveyforge,gottweis2025towards,yamada2025ai,lu2024ai}. This approach holds transformative potential for accelerating the pace of discovery across various scientific domains. By automating tasks such as data analysis, hypothesis generation, experiment design, and result interpretation, automated systems~\citep{yuan2025dolphin, lu2024ai} can efficiently process vast amounts of information and uncover patterns or insights that may be difficult for human researchers to detect.

ASR, while promising, faces significant challenges in generating effective and novel proposals, as well as achieving closed-loop feedback for the experimental validation of these proposals:

\begin{itemize}
    \item First, generating proposals that are both effective and novel is a complicated task. Autonomous systems must identify research gaps and generate hypotheses that are not only innovative but also scientifically valid. This requires balancing creativity and rigor, which is difficult for AI models that rely on existing data and patterns. Additionally, ensuring the novelty of proposals often demands a deep understanding of the broader scientific context, which can be challenging for models limited by the quality and scope of their training data.

    \item Second, implementing closed-loop feedback for end-to-end experimental validation is another major hurdle. Autonomous systems need to design experiments, execute them, analyze results, and iteratively refine their hypotheses in a seamless loop. This requires integration across multiple domains, such as robotics for experiment execution and advanced analytics for result interpretation. Furthermore, real-world experiments often come with unexpected variables and noise, making it challenging for autonomous systems to adapt and learn effectively. Achieving a truly closed-loop system demands robust coordination, adaptability, and the ability to handle uncertainty, which remain technical and conceptual barriers.

\end{itemize}

To further facilitate the advancement of ASR, we propose the InternAgent, an end-to-end auto-research pipeline, which covers four main modules: self-evolving idea generation, human-interactive feedback, idea-to-methodology construction, and multi-round experiment planning and execution. With the help of the Self-Evolving Human-interactive Idea Generation and Idea-to-Methodology Construction, InternAgent can transform a rough proposal into a detailed and easily implementable method, which further increases the success rate of code implementation process and enhances the efficiency of closed-loop experiments. Besides, by leveraging multi-round experiment planning and execution, InternAgent is capable of designing experimental plans and decomposing the experimental process according to the InternAgent-proposed modules, thereby validating the effectiveness of each InternAgent-generated module through experimentation. 

As shown in Fig.~\ref{fig:figure1}, InternAgent has been validated across \textbf{12} scientific research tasks, and we are excited to see that the experimental results demonstrate the significant value of InternAgent in the entire process from hypothesis generation to experimental validation. For instance, in the Reaction Yield Prediction task, the baseline model only achieved a performance of $24.2\%\pm4.2$, while our model improved it to $34.8\%\pm1.1$ in \textbf{just 12 hours}. In contrast, human researchers typically require \textbf{several months} to achieve a similar level of performance improvement. Another example of performance improvement is the Enhancer Activity Prediction task. The baseline model, DeepSTARR, achieved a result of 0.65. By utilizing InternAgent to search relevant domain literature, automatically generate code, and conduct validation, the performance can be improved to 0.79, representing a promising enhancement. In addition, InternAgent also supports complex project-level modifications and debugging, which consist of multiple code files. These results clearly indicate that InternAgent can autonomously generate ideas and design algorithms, effectively reducing the dependence on human effort in scientific research. To facilitate reproducibility, we have open-sourced both the baselines and the codes generated by InternAgent used in all involved scientific research tasks at \texttt{\url{https://github.com/Alpha-Innovator/InternAgent}}.

Furthermore, the contributions of this paper are summarized below:

\begin{itemize}
    \item \textbf{Unified Multi-agent Framework for Diverse Scientific Research Tasks}: 
    We present InternAgent, a unified closed-loop scientific research framework that can automate the entire research cycle, including idea generation, idea-to-methodology transformation, experiment execution, and result feedback. This unified framework can be directly applied to various scientific research scenarios and fields.
    \item \textbf{Interactive Interfaces for Enhanced Cooperative Research}: InternAgent offers interactive interfaces for human-machine collaboration within the idea generation module and across the entire system. By selecting collaboration modes, such as leveraging AI or human experts, it provides evaluations of idea generation effectiveness and facilitates the assessment, reflection, and documentation of experimental results.
    \item \textbf{Comprehensive Experimental Validation and Human Studies}: We conducted extensive human studies centered around InternAgent, including inviting domain experts to evaluate and score the novelty of ideas generated by InternAgent, and comparing the research efficiency between human researchers and InternAgent. These experiments and human studies are crucial for gaining insights into the capabilities of multi-agent systems in conducting scientific research tasks in open-ended environments. We observed many promising phenomena, while also identifying certain technical modules that require improvement.
\end{itemize}

%% file: sec/2_framework.tex
\begin{figure}[t]
\vspace{-5pt}
    \centering
    \includegraphics[width=\linewidth]{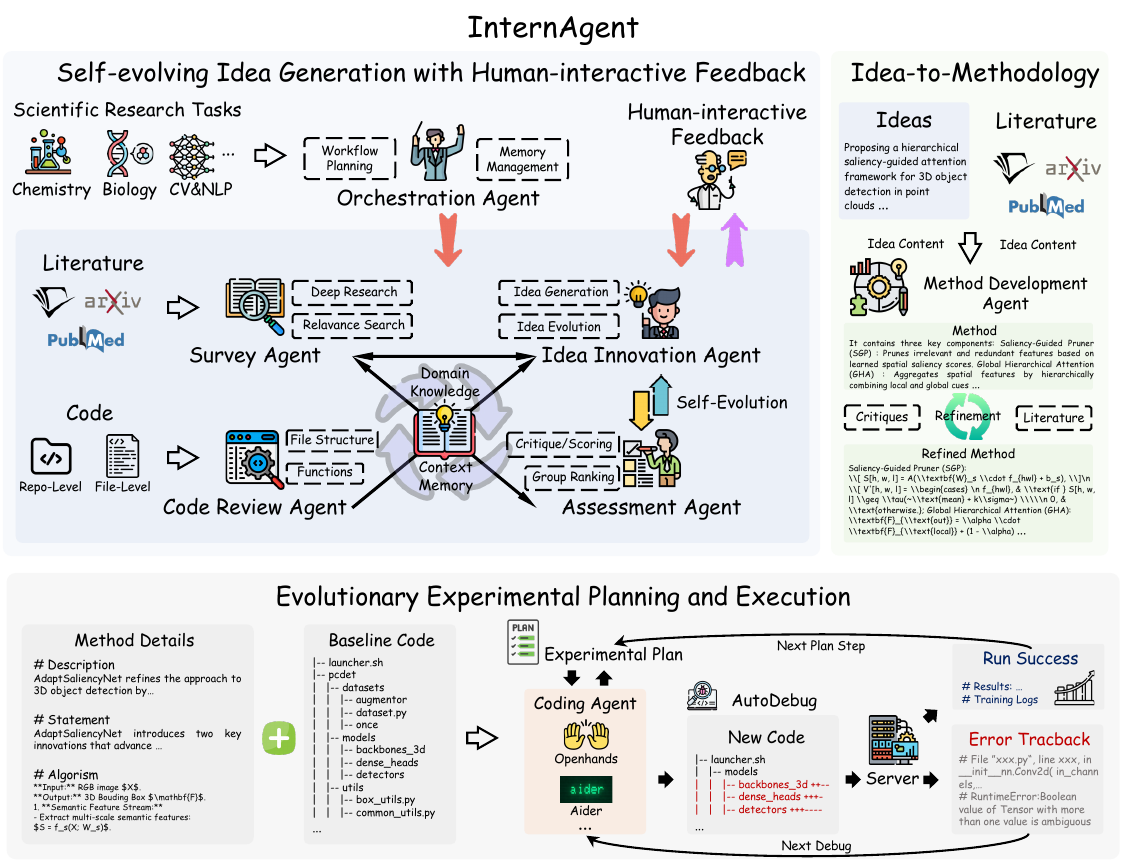}
    \vspace{-10pt}\caption{\textbf{InternAgent} covers three main capabilities: 1) Self-evolving Idea Generation with Human-interactive Feedback, 2) Idea-to-Methodology Construction, and 3) Evolutionary Experimental Planning and Execution.}
    \label{fig:framework}
\end{figure}

\section{InternAgent}
\label{sec:framwork}
As a unified closed-loop multi-agent framework for ASR, InternAgent is designed to facilitate innovative research across diverse scientific domains, as illustrated in Fig.~\ref{fig:framework}. It incorporates three primary capabilities: self-evolving idea generation with human-interactive feedback (Sec.~\ref{sec:idea_gen}), comprehensive idea-to-methodology construction (Sec.~\ref{sec:idea_to_method}), and multi-round automated experiment execution (Sec.~\ref{sec:exp_excu}). Each capability is realized through the collaboration of specialized agents, allowing for seamless integration of different processes to enhance scientific discoveries.

\subsection{Self-Evolving Idea Generation with Human-interactive Feedback}
\label{sec:idea_gen} 
The self-evolving idea generation capability is central to InternAgent, enabling the framework to autonomously generate and refine innovative research ideas. This process involves several specialized agents, each contributing to different stages of idea development and refinement.

\noindent \textbf{Survey Agent.} 
\label{sec:survey_agent}
The Survey Agent is designed to meet the diverse needs of various scientific research tasks by adaptively aligning with user-specified requirements and the necessary depth of detail for exploring existing methodologies. This adaptability is crucial for effectively generating new ideas across different research contexts, and the primary responsibility of the Survey Agent is to search for scientific papers, offering two distinct modes to address the varying needs for depth and breadth in literature research during the scientific discovery process: 1)~\underline{the literature review mode} and 2)~\underline{the deep research mode}.

In the literature review mode, the Survey Agent deconstructs the research task into multiple keyword combinations, enabling a broad search across various academic databases. It collects scientific literature from these sources and evaluates the relevance of each document by analyzing abstracts in relation to the task at hand. Denote the keyword generation process by the function $ P: \mathcal{T} \rightarrow \mathcal{K} $, where $\mathcal{T}$ represents the descriptions of research tasks and $\mathcal{K}$ is the set of generated keyword combinations. The relevance evaluation of each document can be represented by the function:
\begin{equation}
    R: \mathcal{L}_{abs} \times \mathcal{T} \rightarrow [0, 1],
\end{equation}
where $\mathcal{L}_{abs}$ is the abstract of retrieved literature $\mathcal{L}$, and $R(r, t)$ measures the relevance of literature $l$ to the task $t$ as a floating-point score between 0 and 1, with higher scores indicating greater relevance.

In the deep research mode, following the initial literature survey, the Survey Agent downloads and thoroughly examines the full texts of relevant scientific papers. This deeper analysis allows it to generate new keyword combinations, facilitating further rounds of literature exploration. The process of generating new keywords can be denoted by the function:
\begin{equation}
    P: \mathcal{L} \rightarrow \mathcal{K}',
\end{equation}
where $\mathcal{K}'$ is the expanded set of keyword combinations generated from the detailed analysis of full texts. 

By dynamically adjusting its search strategies based on the context of the research stage, the Survey Agent ensures a comprehensive and nuanced understanding of the research landscape. This capability not only supports the generation of innovative ideas but also ensures that the InternAgent framework remains at the cutting edge of scientific discovery.

\noindent \textbf{Code Review Agent.} 
\label{sec:code_review_agent}
The Code Review Agent is crucial for understanding baseline codes for different research tasks, serving as a foundation for innovation by identifying improvements and developing new methodologies. It provides detailed analyses of code structures, dependencies, and functionalities, enabling InternAgent to fully comprehend existing code-bases and identify potential enhancements to advance research objectives. Moreover, the agent's ability to document and summarize complex code-bases ensures efficient navigation and utilization of existing methods. The agent manages two scenarios: 1)~\underline{reviewing user-provided code} or 2)~\underline{searching for relevant code-bases}. For user-uploaded code, it conducts a comprehensive review of the structure, logic, and functionality. Alternatively, in the absence of user-uploaded code, it searches public repositories like GitHub to find relevant code-bases, performing thorough analyses at both the repository and file levels to understand inter-dependencies and assess logic, efficiency, and correctness. Furthermore, the agent employs static code analysis using Python's `ast` module to parse and understand code structure without execution, while the LLM generates human-readable descriptions and summaries, transforming technical details into structured documentation. By using parallel processing with Python's `multiprocessing` module, the agent enhances efficiency and scalability for large code-bases. Overall, the Code Review Agent offers detailed documentation that deepens the understanding of code repositories and supports innovation in scientific research.

\noindent \textbf{Idea Innovation Agent.}
\label{sec:idea_innovation_agent}
The Idea Innovation Agent is an integral part of InternAgent, designed to enhance the creative and iterative processes of scientific research. This agent plays a crucial role by automating the generation and evolution of ideas, thereby addressing the limitations of traditional research works~\citep{yuan2025dolphin, yamada2025ai, lu2024ai}, which often rely on time-consuming manual efforts and are constrained by human cognitive biases. The agent's dual responsibilities, idea generation and idea evolution, are specifically designed to address the diverse needs of various scientific disciplines.

In the context of idea generation, the agent utilizes a general LLM configured with a higher temperature setting. This configuration encourages the generation of more diverse and creative outputs. This enables the agent to identify patterns and insights that might be overlooked in traditional research, generating novel hypotheses and strategies based on task definitions, baseline methods, and current scientific knowledge. The process can be represented by the function:
\begin{equation}
    G: (\mathcal{T}, \mathcal{B}, \mathcal{L}) \rightarrow \mathcal{I},
\end{equation}
where $\mathcal{B}$ denotes analysis of baseline methods and $\mathcal{I}$ is the set of generated ideas. The LLM facilitates the exploration of a broader spectrum of possibilities, accelerating the pace of discovery and innovation by leveraging its comprehensive understanding of language and context.

Idea evolution leverages the capabilities of an LLM to improve existing ideas. The process involves analyzing the content of these ideas, incorporating reflections, which include evaluations of novelty, feasibility, and scientific validity, and integrating insights from related literature. This approach enables the generation of refined and innovative ideas by addressing the inherent limitations of initial concepts. The process can be represented by the same function:
\begin{equation}
    G: (\mathcal{I}, \mathcal{C}, \mathcal{L}) \rightarrow \mathcal{I}',
\end{equation}
where $\mathcal{I}$ is the initial set of ideas, $\mathcal{C}$ denotes the critique, and $\mathcal{I}'$ is the set of evolved ideas. 

Overall, the Idea Innovation Agent enhances scientific ideas into viable and creative solutions by synthesizing and contextualizing information. It critically examines current ideas and employs feedback loops with human experts and other InternAgent agents for continuous improvement. This iterative process balances novelty, feasibility, and ethical considerations, producing impactful and well-rounded ideas.

\noindent \textbf{Assessment Agent.}
\label{sec:assessment_agent}
The Assessment Agent is a vital component of InternAgent, designed to ensure the quality and viability of generated ideas through a rigorous evaluation process. In the rapidly evolving landscape of scientific research, the systematic and objective assessment of ideas is essential. Traditional methods often suffer from subjectivity and lack comprehensive coverage of all relevant dimensions, which can lead to promising ideas being overlooked\citep{qiu2025ai, si2024can}. Therefore, the Assessment Agent addresses these challenges by providing a structured and multidimensional evaluation process, which in turn enhances the reliability and effectiveness of idea selection.

The primary responsibility of the Assessment Agent is to critically evaluate ideas using multidimensional scoring. Each idea is analyzed across key dimensions: coherence, credibility, verifiability, novelty, and alignment. Coherence checks the logical consistency and structure of the idea, while credibility assesses its trustworthiness based on existing knowledge. Verifiability examines the idea's testability through empirical methods. Novelty measures originality, and alignment ensures consistency with research goals.

Moreover, for each dimension, the Assessment Agent provides a detailed evaluation narrative to explain its reasoning. It assigns scores from 0 to 10, which are combined using a weighted summation to produce an overall score for each idea, aiding in the ranking process. By utilizing advanced LLMs, the agent can accurately process and evaluate complex scientific concepts. This capability allows a comprehensive assessment that includes both qualitative and quantitative aspects, ensuring the evaluation is thorough and well-rounded.

Furthermore, the Assessment Agent possesses the ability to ensure diversity among top-ranked ideas. This capability prevents high-scoring ideas from being overly similar or derived from the same original concept. By promoting a varied pool of ideas, the agent encourages the exploration of diverse pathways in the research process. This is crucial for maintaining a balance between innovation and practicality, ensuring that the most promising ideas are both high-quality and distinct from each other.

In summary, by employing LLMs for multidimensional scoring and leveraging its ability to promote diversity among ideas, the Assessment Agent ensures that only the most viable and innovative concepts are selected for further development. This process not only enhances the efficiency of the research cycle but also fosters a more dynamic and diverse research environment.

\noindent \textbf{Human-interactive Feedback.}
\label{sec:human_feedback}
In the context of multi-agent systems, human-interactive feedback is a crucial component for effectively managing and solving complex tasks. This integration of human insights enables agents to navigate dynamic environments more effectively, aligning their outputs with complex user requirements and ensuring practical applicability. 

The human-interactive feedback mechanism of InternAgent is categorized into two primary types: 1)~\underline{feedback directly provided by humans} and 2)~\underline{feedback automatically generated by agent}. Human-provided feedback can address one or multiple ideas, offering insights and critiques that lead to further refinement and adjustment of these ideas based on the feedback received. This iterative process facilitates the continuous improvement of ideas, ensuring they are honed to meet specific objectives and challenges.

For example, in a scenario involving medical image segmentation, an LLM multi-agent system might initially propose a broad idea focused on developing more advanced segmentation algorithms. However, human feedback can refine this idea by directing attention specifically to the medical domain. Human experts can provide insights that encourage the development of adaptive solutions tailored to the unique challenges of medical imaging, such as handling diverse tissue types and ensuring high accuracy in identifying critical structures. This targeted feedback not only sharpens the focus of the idea but also ensures it aligns with the specific needs and priorities of medical research, enhancing its practical applicability and impact.

\noindent \textbf{Orchestration Agent.}
\label{sec:orchestration_agent}
The Orchestration Agent coordinates all other agents within the system, facilitating collaboration by synchronizing tasks and managing data flow. This ensures the process remains efficient, coherent, and aligned with research objectives, allowing the framework to function as an effective research tool. 

Central to the Orchestration Agent's role is designing and managing workflows among agents like the Survey Agent, Code Review Agent, Idea Innovation Agent, and Assessment Agent. It also oversees the timing of human feedback, especially for high-scoring ideas. This requires understanding each agent's capabilities and their interactions to optimize task execution and completion. For example, the Survey Agent conducts adaptive literature exploration, providing insights that the Idea Innovation Agent uses to generate novel hypotheses. The Orchestration Agent ensures these findings are communicated effectively. Similarly, it synchronizes the Code Review Agent's analyses to enhance idea evaluation and development. Furthermore, the Orchestration Agent manages the Assessment Agent’s evaluation process, ensuring timely and relevant outputs. This helps guide the development of diverse top ideas. Additionally, it determines optimal points for human feedback, integrating expert insights after identifying high-scoring ideas to refine and adapt them, aligning outputs with user requirements.

\begin{figure}[t]
\vspace{-10pt}
    \centering
    \includegraphics[width=\linewidth]{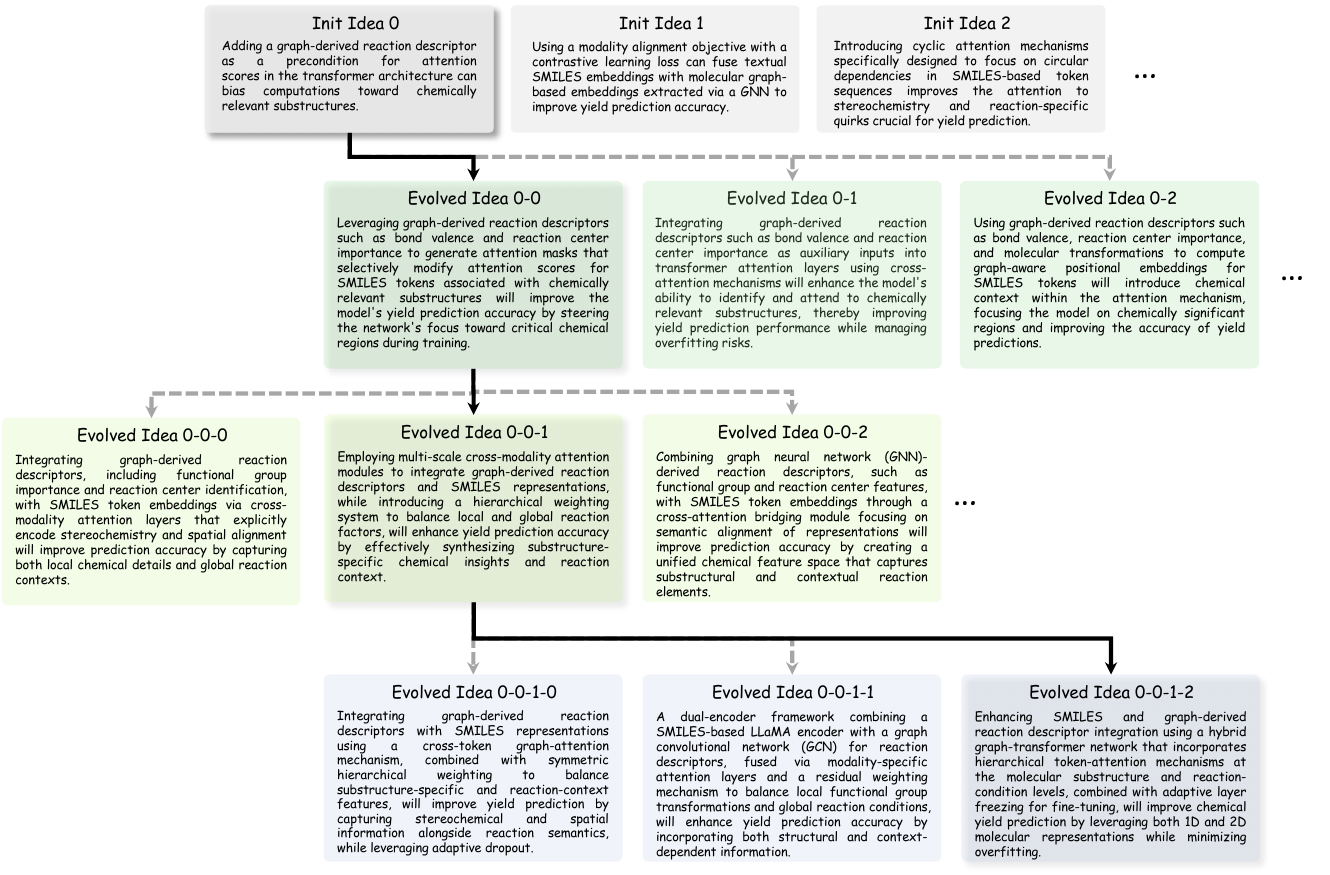}
    \vspace{-10pt}\caption{InternAgent Self-evolutionary path of ideas for reaction yield prediction task.}
    \label{fig:idea_evol}
\end{figure}

In summary, as illustrated in Fig.~\ref{fig:idea_evol}, by managing multi-agent collaboration and integrating human feedback, the Orchestration Agent enables InternAgent to operate as a cohesive and innovative research tool, driving scientific discovery forward.

\subsection{Comprehensive Idea-to-Methodology Construction}
\label{sec:idea_to_method}

The idea-to-methodology construction process systematically bridges the gap between concise research ideas and concrete, implementable methodologies, ensuring that the AI-generated ideas could be realized and their validity verified. This process is orchestrated by the Methodology Development Agent, which collaborates closely with other agents and integrates both automated processes and human-interactive feedback loops to ensure that methodological development is rigorous, traceable, and practically relevant. Specifically, to develop a comprehensive method corresponding to the concise research idea, the Method Development Agent possesses two core capabilities: 1)~\underline{Methodology Initialization}: which involves constructing the basic structure and content of a method by integrating the idea with baseline codes and the methodology content of relevant literature; 2)~\underline{Methodology Refinement}: which iteratively enhances the basic method structure for the purpose of rigor and completeness, ensuring a more detailed and robust methodology.

\subsubsection{Methodology Initialization}

To convert concise research ideas into detailed methodological frameworks, the Method Development Agent uses its Methodology Initialization capability. The process begins by extracting core objectives and hypotheses from research ideas, identifying key variables, and understanding their interrelationships to construct a coherent framework. The agent uses multiple resources: task descriptions $\mathcal{T}$ provide context and constraints; baseline implementations $\mathcal{B}$ offer adaptable methods; and relevant literature $\mathcal{L}$ integrates existing knowledge and ensures that the framework aligns with current research.

By formalizing mechanisms that require empirical investigation, the agent details processes and conditions for conducting research and specifies methods for data collection and analysis. The outcome is a methodological framework that is both theoretically sound and practically executable. The transformation function is represented as:
\begin{equation}
    T: \mathcal{I} \times \mathcal{T} \times \mathcal{B} \times \mathcal{L} \rightarrow \mathcal{M},
\end{equation}
where $\mathcal{I}$ denotes research ideas, $\mathcal{T}$ includes task descriptions, $\mathcal{B}$ represents baseline methods, $\mathcal{L}$ is the literature corpus, and $\mathcal{M}$ is the resulting methodological framework.
Overall, through Methodology Initialization, the Method Development Agent effectively turns initial ideas into detailed, actionable methods, ready for further refinement.

\subsubsection{Methodology Refinement}

After the initialization, the Methodology Development Agent leverages its refinement capability to critically evaluate and iteratively improve the methodological framework. The agent conducts a comprehensive analysis of the initial methodology $\mathcal{M}$, incorporating structured critiques $\mathcal{C}$, which include both automated assessments and expert human feedback. Additionally, it synthesizes insights from the latest scientific literature $\mathcal{L}$. The refinement process is formally defined as:
\begin{equation}
    R: \mathcal{M} \times \mathcal{C} \times \mathcal{L} \rightarrow \mathcal{M}',
\end{equation}
where $\mathcal{M}$ represents the initial methodology, $\mathcal{C}$ denotes the critique space, potentially including human feedback and automated assessments, $\mathcal{L}$ is the literature corpus, and $\mathcal{M}'$ is the refined methodological framework.

During both initialization and refinement, the Methodology Development Agent collaborates closely with other agents, such as the Assessment Agent for multidimensional evaluation and the Orchestration Agent for workflow coordination. This collaboration ensures that each methodological step benefits from comprehensive feedback and current domain knowledge. The integrated, multi-agent approach guarantees that the transformation from idea to methodology is systematic and adaptable, supporting the continuous evolution and optimization of scientific research within the InternAgent framework.

\subsection{Evolutionary Experimental Planning and Execution} 
\label{sec:exp_excu}

\subsubsection{Exception-Guided Debugging Framework} Converting theoretical concepts into functional code is challenging. To this end, we developed an exception-guided debugging framework that systematically converts abstract methodological text descriptions into executable implementation codes. This framework operates by systematically capturing runtime exceptions during execution attempts, analyzing error contexts, and formulating targeted fixes through reasoning of the large language model.

Our coder module employs a dual-strategy approach according to the complexity of given baseline code. For single-file or limited-scope implementation tasks, we use the Aider coding assistant~\citep{aider}, which facilitates localized code modifications with minimal overhead. For complex repository-level codes requiring comprehensive structural understanding across different functions, we deploy OpenHands framework~\citep{openhands}, which enables thorough codebase analysis and coordinated multi-file modifications while maintaining the integrity of the overall code architecture.

Once the initial code implementation is completed, the framework transitions to a systematic debugging phase to ensure functionality and robustness. The debugging process follows a systematic cycle: (1) execution attempt, (2) exception capture and traceback analysis, (3) contextual code structure understanding, (4) debugging strategy formulation, and (5) targeted implementation. This cycle continues iteratively until successful execution or reaching a predefined iteration threshold.

\subsubsection{Experimental Planning and Adaptive Evolution} 
\label{exp_evo}

After establishing basic functionality through debugging, we transition to implementation planning focused on identifying critical structures and integration points. Our planning process first determines which core modules require modification, then develops a step-by-step implementation strategy with clear priorities and dependencies.

Implementation planning operates at multiple abstraction levels: architectural modifications for methodological alignment, algorithmic transformations for core functionality, and optimization adjustments for performance characteristics. This approach aims to provide structure when implementing methodological improvements across interconnected components in AI systems, which helps guide development efforts.

Rather than employing a single-pass implementation strategy, we designed an adaptive evolution approach for our implementation process. This approach involves structured iterations where each implementation attempt is followed by performance assessment and potential refinement. We maintain records of implementation decisions across iterations, which helps track changes and their corresponding effects. This directed adaptation process enables the gradual refinement of complex implementations based on empirical results rather than theoretical assumptions alone.

%% file: sec/3_quantitative.tex
\section{Experiments}
\label{sec:exp}

In this section, we evaluate the effectiveness of InternAgent in conducting autonomous research and accelerating scientific discovery. We begin by providing a brief overview of the selected multi-domain tasks and detailing the experimental implementation in Sec.~\ref{sec:experimental_setup}. Subsequently, we present the quantitative results across various tasks in Sec.~\ref{sec:experimental_results} and conduct an analysis of the different modules within InternAgent in Sec.~\ref{sec:insightful_analyses}.

\subsection{Experimental Setup}
\label{sec:experimental_setup}

\subsubsection{Task Description}

We select 12 distinct tasks to demonstrate InternAgent's capability in conduct Autonomous Scientific Research (ASR). These tasks span multiple modalities, including science (\textit{e.g.}, reaction yield prediction, molecular dynamics), time series (\textit{e.g.}, time series forecasting), natural language (\textit{e.g.}, sentiment classification), image (\textit{e.g.}, semantic segmentation), and point cloud (\textit{e.g.}, 3D object detection), which cover both discriminative and generative tasks. We believe that experiments ranging from fundamental tasks to complex multi-modal tasks can comprehensively illustrate the effectiveness of InternAgent. Below, we detail the datasets, the base code repositories, and the experimental settings for each task.

\begin{itemize}
    \item \textbf{Reaction Yield Prediction (AutoRYP).} We conduct experiments on the widely-used Suzuki-Miyaura reaction dataset~\citep{perera2018platform}, which contains 5,760 reaction data. Each data point includes structured chemical reaction information, such as reactants, products, reaction types, reaction conditions (solvent, catalyst, ligand, and base), functional group, and yield values. We use the LoRA-finetuned LLaMA3-8B as our baseline, an embedding model that converts chemical reaction texts into high-dimensional vector representations, which are subsequently fed into a fully connected prediction network predictor to perform chemical yield prediction.
    \item  \textbf{Molecular Dynamics (AutoMD).} We conduct experiments on the widely-used MD17 dataset~\citep{chmiela2017machine}, which contains energy and force calculation results for seven small organic molecules: aspirin, ethanol, malonaldehyde, naphthalene, salicylic acid, toluene, and uracil. We use VisNet~\citep{wang2024enhancing} as our baseline, an equivariant geometry-enhanced graph neural network that achieves excellent chemical property prediction.
    \item  \textbf{Power Flow Estimation (AutoPower).} We conduct experiments on the IEEE 39-Bus dataset~\citep{zimmerman2010matpower}, which is a medium-scale benchmark based on the New England power system, comprising 39 buses, 10 synchronous generators, 19 load buses and 46 transmission lines, and providing AC power flow snapshots under a variety of load conditions. We use SenseFlow~\citep{zhao2024senseflow} as our baseline, a novel physics-informed, self-ensembling power flow estimation model that has demonstrated state-of-the-art accuracy on standard IEEE test systems  consistently outperforming both traditional state-estimation techniques and recent data-driven approaches in voltage and power‐flow recovery tasks.
    \item  \textbf{Time Series Forecasting (AutoTSF).} We conduct experiments on the ETTh1 dataset, which is a 1-hour-level subset of the Electricity Transformer Temperature (ETT) benchmark~\citep{haoyietal-informer-2021}. This dataset comprises two years of hourly multivariate time series, including the target oil temperature and six power-load covariates, collected from transformer stations in two Chinese counties. We use DLinear~\citep{zeng2023transformers} as our baseline, an MLP-based forecasting model that decomposes each series into trend and seasonality and employs simple linear layers, outperforming Transformer-based methods on multiple time series benchmarks.  We report the average results of 96, 192, 336, and 720 prediction length.
    \item  \textbf{Transcription Prediction for Perturbation Response (AutoTPPR).} We conduct experiments on the Perturb-seq dataset~\citep{norman2019exploring}, which contains single-cell gene expression data measuring the transcriptional responses of cells to various perturbations.  We use GEARS (Generative Energy-based Autoencoder for scRNA-seq)~\citep{roohani2024predicting} as our baseline, a framework based on Graph Neural Networks (GNNs) and Multi-Layer Perceptrons (MLPs), designed to learn joint representations of single-cell multi-omics data.
    \item \textbf{Enhancer Activity Prediction (AutoEAP).} We conduct experiments on the UMI-STARR-seq dataset~\citep{arnold2013genome}, which contains genome-wide, high-resolution quantitative activity maps of developmental and housekeeping enhancers in Drosophila S2 cells. We use DeepSTARR~\citep{de2022deepstarr} as our baseline, a deep learning model that excels at quantitatively predicting enhancer activity from DNA sequences.
    \item \textbf{Sentiment Analysis (AutoSenCls).} We conduct experiments on the Stanford Sentiment Treebank (SST-2) dataset~\citep{socher2013recursive}, a binary sentiment classification dataset consisting of movie reviews with approximately 67,000 training samples. We use BERT-base~\citep{devlin2019bert} as our baseline, a Transformer-based pre-trained language model that has shown excellent performance on various NLP tasks.  

    \item \textbf{2D Image Classification (Auto2DCls).} We conduct experiments on the widely-used CIFAR-100 dataset~\citep{krizhevsky2009cifar}, which contains 60,000 32×32 color images in 100 classes, with 500 training images and 100 testing images per class. We use Wide Residual Networks (WRN)~\citep{zagoruyko2016wrn} as our baseline, which improves performance by increasing the width rather than the depth of convolutional networks.
    \item \textbf{3D Point Cloud Classification (Auto3DCls).} We conduct experiments on the ModelNet40 dataset~\citep{wu2015modelnet}, which contains 12,311 CAD models across 40 common object categories and is widely used for 3D shape classification tasks. We use PointNet~\citep{qi2017pointnet} as our baseline, a pioneering deep learning architecture that directly processes point cloud data.
    \item \textbf{2D Semantic Segmentation (Auto2DSeg).} We conduct experiments on the widely-used Pascal VOC 2012 dataset~\citep{pascal-voc-2012}, which includes 20 object classes and a background class for semantic segmentation tasks. The dataset contains 1,464 images for training and 1,449 for validation. We use DeepLabV3Plus~\citep{chen2018encoder} as our baseline method, which enhances segmentation performance by employing atrous convolution and a more refined encoder-decoder structure to capture multiscale contextual information effectively.
    \item \textbf{3D Point Cloud Autonomous Driving (AutoPCDet).} We conduct experiments on the widely-used dataset ONCE~\citep{mao2021once} and use CenterPoint~\citep{yin2021center} as our baseline. Our code is based on OpenPCDet~\citep{openpcdet2020} and we filter out all code irrelevant to the baseline model to avoid knowledge leakage.
\item \textbf{Large Vision-Language Model Fine-tuning (AutoVLM).}We conduct experiments on filtered geometry subset of the URSA dataset~\citep{luo2025ursa}, comprising manually curated multimodal QA pairs and CoT process. Natural images were excluded, and data were downsampled to control experimental budgets, enabling training completion within 20 hours on 8 A800 GPUs.We use LLaVA-Onevision~\citep{li2024llava} as our baseline, a robust multimodal alignment framework using a simple MLP to align visual encoders with LLMs, forming an effective LMM with strong scalability on vision-language tasks. We take SigLIP~\citep{zhai2023sigmoid} and Qwen2.5-Math-7B-Instruct\citep{yang2409qwen2} as the visual and language modules, respectively.
\end{itemize}

\subsubsection{Evaluation Metric}
Since our InternAgent has been validated across a wide range of scientific research fields, the evaluation metrics used for tasks in each field are not consistent. Therefore, in this part, we provide a detailed introduction to the evaluation metrics used for each scientific research task.

\begin{itemize}
    \item \textbf{AutoRYP.} For Reaction Yield Prediction, we evaluate model performance using the coefficient of determination (R²), which quantifies the proportion of variance in the actual reaction yields that is predictable from the model's predictions.
    \item \textbf{AutoMD.} Our method is evaluated on the MD17 dataset, a molecular chemical property prediction task. The performance is measured using Force-MAE, representing the mean absolute error between the true and predicted forces of molecules.
    \item  \textbf{AutoPower.} For Power Flow Estimation, we use Root Mean Square Error (RMSE) on PQ node to evaluate the estimation performance on IEEE 39-Bus datasets, representing the root mean square error between the true and predicted voltage magnitudes and phase angles.

    \item  \textbf{AutoTSF.}  For Time Series Forecasting, we use Mean Absolute Error (MAE) to evaluate the prediction performance on ETTh1 dataset. The performance is calculated by taking the average of the four prediction steps of \{96, 192, 336, 720\}.
    
    \item  \textbf{AutoTPPR.} For Transcription Prediction for Perturbation Response, we employ the Top 20 DE MSE as the evaluation metric, calculating the mean squared error between the predicted and actual expression levels of the top 20 most differentially expressed genes under each perturbation condition.
    
    \item \textbf{AutoEAP.} For Enhancer Activity Prediction, we use Housekeeper Pearson Correlation Coefficient (HK-PCC) as the metric, which quantifies the correlation between the true enhancer activities and the predicted values.
    \item \textbf{AutoSenCls.} We evaluate our method on the SST-2 dataset, which is a binary sentiment classification task. The performance is measured using accuracy (Acc), which represents the percentage of correctly classified samples.
    \item \textbf{Auto2DCls.} For 2D image classification, we conduct experiments on CIFAR-100 dataset, which contains 100 classes. The performance is measured using classification accuracy (Acc), representing the percentage of correctly classified images.
    \item \textbf{Auto3DCls.} For the task of 3D point cloud classification, we use the widely adopted ModelNet40 benchmark, which comprises 40 distinct object categories. We report the Overall Accuracy (OA) as our primary evaluation metric, which calculates the proportion of correctly classified instances in the entire test set.
    \item \textbf{Auto2DSeg.} For 2D semantic segmentation, we conduct experiments on the Pascal VOC 2012 dataset, which includes 20 object classes and a background class. The performance is measured using the mean Intersection over Union (MIoU), which quantifies the average overlap between the predicted segmentation and the ground truth across all classes, providing a comprehensive assessment of the model's segmentation accuracy.
    \item \textbf{AutoPCD.}  Following ONCE official evaluation metric, we merge the car, bus and truck class into a super-class (\textit{i.e.}, vehicle). $\text{AP}_{3D}$ is used to evaluate the performance of the ONCE dataset, we report Mean average precision (mAP) which is the average of the scores of the three categories.
    \item \textbf{AutoVLM.} We evaluated our model on the geometry subset of MathVista~\citep{lu2023mathvista}, a widely adopted multimodal mathematical benchmark. Model's answers to questions were extracted using GPT-4o and compared against the ground truth to calculate accuracy.
\end{itemize}

\subsubsection{Implementation Details}
\label{sec:implementation_detail}

In the self-evolving idea generation process, the survey agent, code review agent, generation agent, self-evolving agent, and orchestration agent are based on GPT-4o~\citep{hurst2024gpt4o}. The survey agent searches and reviews 50 papers to provide domain knowledge for the subsequent idea generation agent, and then the idea generation agent generates 15 ideas. The self-evolving agent evolves each idea into 3 ideas and then selects the top 5 ideas for the next evolving process until the maximum number of evolutions (\textit{i.e.}, 4) is reached. In the idea-to-methodology process, each idea is initialized and refined once by the method development agent. In the evolutionary experimental planning and execution process. We use Claude-3.7-Sonnet to generate codes and debug. We set the max debug attempt to 4. The max run number is set to 5 for Aider~\citep{aider} and 3 for OpenHands~\citep{wang2024openhands}.

\subsection{Experimental Results}
\label{sec:experimental_results}

\begin{table}[tbp]
\centering
\renewcommand{\arraystretch}{1.2}
\caption{Performance comparison across six types of scientific research tasks. We conduct experiments using 10 InternAgent-generated ideas for each task.}
\vspace{-6pt}
\label{tab:performance_comparison_sci}
\resizebox{0.97\textwidth}{!}{
\begin{tabular}{l@{\hspace{0.5em}}cccccc}
\toprule
& \multicolumn{6}{c}{\textbf{Tasks}} \\
\cmidrule(lr){2-7}
\textbf{Method} & \textbf{AutoRYP} & \textbf{AutoMD} & \textbf{AutoPower} & \textbf{AutoTSF} & \textbf{AutoTPPR} & \textbf{AutoEAP} \\
\cmidrule(lr){2-7}
& \textbf{$R^2$} & \textbf{Forces-MAE} & \textbf{RMSE} & \textbf{MAE} & \textbf{MSE} & \textbf{HK-PCC}  \\
\midrule
\multicolumn{7}{c}{\textbf{Max Performance}} \\
\midrule
Baseline & 27.6 & 0.158 & 0.00473 & 0.4382 & 0.197 & 0.65 \\
Dolphin & 31.8 (+4.2) & 0.152 & 0.00455 & 0.4627 & 0.173 & 0.76  \\
InternAgent & \textbf{35.4 (+7.8)} &  \textbf{0.148} &  \textbf{0.00426} &  \textbf{0.4331} & \textbf{0.146} & \textbf{0.79} \\
\midrule
\multicolumn{7}{c}{\textbf{Average Performance}} \\
\midrule
Baseline & 27.6 & 0.158 & 0.00473 & 0.4382 & 0.197 & 0.65 \\
Dolphin & 31.3 (+3.7) & 0.155 & 0.00459 & - & 0.179 & 0.73  \\
InternAgent & \textbf{33.5 (+5.9)} & \textbf{0.152} & \textbf{0.00447} & \textbf{0.4346} & \textbf{0.170} & \textbf{0.77}  \\
\bottomrule
\end{tabular}
}
\end{table}

\begin{table}[t]
\centering
\renewcommand{\arraystretch}{1.2}
\caption{Performance comparison for six types of scientific research tasks. We conduct experiments using 10 InternAgent-generated ideas for each task, where baseline codes for Auto2DSeg, AutoPCDet, and AutoVLM are project-level, consisting of multiple code files with complex call relation between functions. Therefore, the coder in Dolphin~\citep{yuan2025dolphin} does not support modifying this type of baseline codes.}
\vspace{-6pt}
\label{tab:performance_comparison_ai}
\resizebox{\textwidth}{!}{
\begin{tabular}{l@{\hspace{0.5em}}ccccccc}
\toprule
& \multicolumn{6}{c}{\textbf{Tasks}} \\
\cmidrule(lr){2-7}
\textbf{Method} &  \textbf{AutoSenCls}  & \textbf{Auto2DCls} &  \textbf{Auto3DCls} &  \textbf{Auto2DSeg} & \textbf{AutoPCDet} & \textbf{AutoVLM} \\
\cmidrule(lr){2-7}
&   \textbf{Acc} & \textbf{Top-1 Acc} & \textbf{OA} & \textbf{mIoU} & \textbf{mAP} & \textbf{QA} \\
\midrule
\multicolumn{7}{c}{\textbf{Max Performance}} \\
\midrule
Baseline & 91.0  & 81.2 & 91.0  & 78.8 & 65.0 & 67.1\\
\textsc{Dolphin} & 92.5 (+1.5) & 82.0 (+0.8) & 93.9 (+2.9)  & - & -  & -\\
InternAgent & \textbf{93.5 (+2.5)} & \textbf{83.3 (+2.1)} & \textbf{95.5 (+4.5)} & \textbf{81.0 (+2.2)} & \textbf{65.9 (+0.9)} & \textbf{67.6 (+0.5)} \\
\midrule
\multicolumn{7}{c}{\textbf{Average Performance}} \\
\midrule
Baseline & 91.0  & 81.2 & 91.0 & 78.8 & 65.0 & 67.1\\
\textsc{Dolphin} &  91.8 (+0.8) & 81.8 (+0.6) & 92.0 (+1.0)  &  - & -  & -\\
InternAgent & \textbf{92.5 (+1.5)}  & \textbf{82.2 (+1.0)} & \textbf{93.4 (+2.4)} & \textbf{80.1 (+1.3)} & \textbf{65.7 (+0.7)} & \textbf{67.6 (+0.5)} \\
\bottomrule
\end{tabular}
}
\end{table}

\begin{table}[tbp]
\centering
\renewcommand{\arraystretch}{1.2}
\caption{Experiments statistics across different tasks. Each cell shows the number of ideas that improved performance, the number of ideas that successfully ran, and the total number of ideas tested (format: improved / successful / tested). For all the tasks, we conduct experiments with 10 ideas.}
\vspace{-6pt}
\label{tab:idea_statistics_sci}
\resizebox{0.90\textwidth}{!}{
\begin{tabular}{l|cccccc}
\toprule
& \multicolumn{6}{c}{\textbf{Research Task}} \\
\cmidrule(lr){2-7}
\textbf{Method} & \textbf{AutoRYP} & \textbf{AutoMD} & \textbf{AutoPower} & \textbf{AutoTSF}  & \textbf{AutoTPPR} & \textbf{AutoEAP} \\
\midrule
Dolphin & 2/3/10 & 2/4/10 & 2/4/10 & 0/3/10& 2/3/10 & 2/4/10 \\
InternAgent & 4/6/10 & 4/8/10 & 5/6/10 & 3/7/10 & 5/5/10 & 8/8/10 \\
\bottomrule
\end{tabular}
}
\end{table}

\begin{table}[t]
\centering
\renewcommand{\arraystretch}{1.2}
\caption{Experiments statistics across different tasks. Each cell shows the number of ideas that improved performance, the number of ideas that successfully ran, and the total number of ideas tested (format: improved / successful / tested). For all the tasks, we conduct experiments with 10 ideas.}
\vspace{-6pt}
\label{tab:idea_statistics_ai}
\resizebox{0.90\textwidth}{!}{
\begin{tabular}{l|cccccc}
\toprule
& \multicolumn{6}{c}{\textbf{Research Task}} \\
\cmidrule(lr){2-7}
\textbf{Method}  & \textbf{Auto2DCls} & \textbf{Auto3DCls} & \textbf{AutoSenCls}  & \textbf{Auto2DSeg} & \textbf{AutoPCDet} & \textbf{AutoVLM} \\
\midrule
Dolphin & 2/4/10 & 2/5/10 & 4/7/10 & - & - & - \\
InternAgent & 5/7/10 & 3/6/10 & 9/9/10 & 6/9/10 & 2/5/10 & 1/5/10 \\
\bottomrule
\end{tabular}
}
\end{table}

To comprehensively evaluate the effectiveness of InternAgent in accelerating scientific discovery, we first provide quantitative experimental results as shown in Tab.~\ref{tab:performance_comparison_sci}, Tab.~\ref{tab:performance_comparison_ai}, Tab.~\ref{tab:idea_statistics_sci}, and Tab.~\ref{tab:idea_statistics_ai}. Extensive results demonstrate that InternAgent excels in the following aspects: 

\begin{table}[t]
\centering
\renewcommand{\arraystretch}{1.2}
\caption{Computational and financial cost analysis for all tasks. Training time is measured using A100 GPU hours, while idea generation and code debugging costs are measured in USD using gpt-4o and claude-3.7-sonnet models respectively.}
\vspace{-6pt}
\label{tab:cost_analysis_sci}
\resizebox{\textwidth}{!}{
\begin{tabular}{l|cccccc}
\toprule
\textbf{Cost Metric} & \textbf{AutoRYP} & \textbf{AutoMD} & \textbf{AutoPower} & \textbf{AutoTSF} & \textbf{AutoTPPR}  & \textbf{AutoEAP} \\
\midrule
Training time (A100 hours) & 6.0 & 10.0 & 5.0 & 0.1 & 1.0 & 1.0\\
Idea-Gen cost (gpt-4o) (\$) & 0.6 & 0.6 & 0.6 & 0.6 &  0.6 &  0.6\\
Coder-Debug cost (claude-3.7-sonnet) (\$) & 0.7 & 0.5 & 1.0& 0.4 & 0.9 & 0.6\\
\bottomrule
\end{tabular}
}
\end{table}

\begin{itemize}
    \item \textbf{Outperforming existing auto-research systems on multiple tasks.} We first compare InternAgent with existing auto-research system (\textit{i.e.}, \textsc{Dolphin}~\citep{yuan2025dolphin}) on single-file tasks. Tab.~\ref{tab:performance_comparison_sci} and Tab.~\ref{tab:performance_comparison_ai} show the max performance and average performance (\textit{i.e.}, the average performance across experiments with performance gains) achieved by InternAgent and \textsc{Dolphin}.  It can be observed that InternAgent consistently improves the performance compared to the baseline and outperforms \textsc{Dolphin} across all tasks including both generative and discriminative tasks. This suggests that  InternAgent can generate better ideas on each specific domain benefiting from the self-evolving idea generation process and automatically implement them. For example, in AutoRYP, methods proposed by InternAgent can largely outperform those proposed by \textsc{Dolphin} (\textit{i.e.}, +3.6 on max performance). We highlight that InternAgent can achieve SoTA performance on some tasks such as 3D point cloud classification (\textit{i.e.}, 95.5\% overall accuracy without pre-training achieved by InternAgent compared to 95.3\% overall accuracy with pre-training achieved by human experts).

    Besides, Tab.~\ref{tab:idea_statistics_sci} and Tab.~\ref{tab:idea_statistics_ai} report the percentage of experiments with performance gains and executable experiments out of the total number of experiments. First, results show that even on complex tasks such as AutoPCDet (\textit{i.e.}, 50\%) and Auto2DSeg (\textit{i.e.}, 90\%), InternAgent can still ensure a reasonable execution success rate which is due to the carefully designed idea-to-methodology process, enabling the coder to auto-implement based on detailed methodologies. Second, InternAgent demonstrates a higher performance improvement rate compared to \textsc{Dolphin}. This improvement is mainly attributed to the idea-to-methodology feature of InternAgent, which enables the concretization of high-level ideas. Additionally, through the process of multi-round experimental planning and execution, the submodules of the AI-generated methodology are progressively integrated into the baseline code.
    
    \item \textbf{Covering a wide range of tasks including the scientific research tasks and AI tasks.} Further, InternAgent exhibits strong generalization capability across a wide range of tasks, enabling it to handle tasks from the AI domain (\textit{e.g.}, Auto2DSeg) to the scientific domain (\textit{e.g.}, AutoMD). As shown in Tab.~\ref{tab:performance_comparison_sci} and Tab.~\ref{tab:performance_comparison_ai}, InternAgent can support 12 different tasks ranging from simple classification tasks to complex multimodal and cross-disciplinary tasks. This is because the survey agent in InternAgent can auto-search task-related literature on academic websites such as arXiv and review the literature to understand each task. Besides, InternAgent is highly extensible, as it can support new tasks with just a task description and reference codes. This capability not only assists AI researchers in automatically updating algorithms, but also empowers researchers in scientific domains to utilize AI tools at a lower cost, thereby accelerating the pace of scientific discovery.
    
    \item \textbf{Support repo-level experiments.} Most of existing auto-research systems such as \textsc{Dolphin}~\citep{yuan2025dolphin} only support single-file experiments. On more complex tasks, researchers are required to manually consolidate complex task codes into a single file, which is highly time-consuming and limits their ability to conduct experiments on complex tasks. In contrast, InternAgent can support repo-level tasks such as AutoPCDet, AutoVLM, AutoTPPR, and so on, and achieve better performance on these repo-level tasks compared to their baselines. For example, on Auto2DSeg, InternAgent pipeline can improve the DeepLabV3Plus baseline~\citep{chen2018encoder}  from the original 78.80\% to 81.0\%. This is attributed to the detailed methodology, code comprehension achieved by the code review agent, and the auto-exploration ability of the coder agent.

\end{itemize}


\textbf{Runtime Statistics.} We further provide the runtime statistics of InternAgent on all 12 tasks including the training costs (\textit{i.e.}, GPU hours) and monetary costs in the idea generation stage (including self-evolving idea generation and idea-to-methodology) and code execution and debug stage. As shown in Tab.~\ref{tab:cost_analysis_sci} and Tab.~\ref{tab:cost_analysis_ai}. As mentioned in Sec.~\ref{sec:implementation_detail}, we select top 5 ideas in each idea generation process and then generate detailed methodology for the selected ideas. Therefore, we report the average cost of 5 ideas as the idea generation cost. It can be seen that the idea generation cost of each idea is about \$0.6 using GPT-4o which is cost-efficient. The coder-debug cost denote the cost of each run, for example, if running for 5 times for a single idea as mentioned in Sec.~\ref{sec:exp_excu}, we calculate the average cost of 5 runs. It can be seen from the table that the coder-debug cost varies between the file-level codes and repo-level codes and repo-level codes generally need more cost for high complexity of codes. For example, for single-file code such as Auto2DCls, the cost is below \$1 for each run and for more complex AutoPCDet, the cost is about \$1.2 using claude-3.7-sonnet. Generally, InternAgent is a cost-efficient auto-research framework that can generate ideas and execute codes at a reasonable cost.

\begin{table}[t]
\centering
\renewcommand{\arraystretch}{1.2}
\caption{Computational and financial cost analysis for all tasks. Training time is measured using A100 GPU hours, while idea generation and code debugging costs are measured in USD using gpt-4o and claude-3.7-sonnet models respectively.}
\vspace{-6pt}
\label{tab:cost_analysis_ai}
\resizebox{\textwidth}{!}{
\begin{tabular}{l|cccccc}
\toprule
\textbf{Cost Metric} & \textbf{Auto2DCls} & \textbf{Auto3DCls} & \textbf{AutoSenCls} & \textbf{Auto2DSeg} & \textbf{AutoPCDet} & \textbf{AutoVLM} \\
\midrule
Training time (A100 hours) & 2.0 & 0.8 & 0.3 & 30.0 & 9.0 & 192.0 \\
Idea-Gen cost (gpt-4o) (\$) & 0.6 & 0.6 & 0.6 & 0.6 & 0.6 & 0.6 \\
Coder-Debug cost (claude-3.7-sonnet) (\$) & 0.7 & 0.6 & 0.7 & 1.1 & 1.2 & 1.0\\
\bottomrule
\end{tabular}
}
\end{table}

\begin{table}[t]
\centering
\renewcommand{\arraystretch}{1.2}
\caption{To compare the performance of the baseline and InternAgent-generated code, we adopted a few-shot training setup for the yield prediction task. Due to the large variance in experimental results under this setting, we report the outcomes of 5 independent repeated experiments.}
\vspace{-6pt}
\label{tab:multi_dimension}
\resizebox{\textwidth}{!}{
\begin{tabular}{l|cccccc}
\toprule
\textbf{Epoch=300} & \textbf{Repeat=1} & \textbf{Repeat=2} & \textbf{Repeat=3} & \textbf{Repeat=4} & \textbf{Repeat=5} & \textbf{AVG/VAR} \\
\midrule
Baseline (train-set=60) & 20.0 & 26.2 & 27.6 & 26.6 & 20.1 & 24.2$\pm$4.2 \\
GAT (ours, train-set=60) & 34.7 & 34.8 & 33.9 & 32.7 & 34.2 & 34.1$\pm$1.4 \\
ADAGT (ours, train-set=60) & 35.4 & 35.2 & 34.5 & 35.2 & 33.7 & \textbf{34.8$\pm$1.1} \\ \midrule
Baseline (train-set=100) & 38.8 & 30.6 & 34.8 & 39.0 & 34.5 & 35.5$\pm$4.9 \\
GAT (ours, train-set=100) & 36.9 & 39.1 & 34.4 & 41.4 & 35.0 & 37.4$\pm$4.0 \\
ADAGT (ours, train-set=100) & 38.5 & 38.0 & 38.6  & 37.9 & 40.4 & \textbf{38.7$\pm$1.7} \\
\bottomrule
\end{tabular}
}
\end{table}

\begin{table}[t]
\centering
\caption{Ablation Study on Adaptive Evolution (AE). Ideas (i/s/t) shows the number of ideas that improved performance, the number of ideas that successfully ran, and the total number of ideas tested (format: improved / successful / tested).}
\vspace{-8pt}
\label{tab:ablation_study_combined}
\renewcommand{\arraystretch}{1.1}
\resizebox{\textwidth}{!}{%
\begin{tabular}{l|cc|c|cc|c|cc|c}
\toprule
\multirow{2}{*}{\textbf{Method}} &  \multicolumn{3}{c|}{\textbf{AutoRYP}} & \multicolumn{3}{c|}{\textbf{Auto2DCls}} & \multicolumn{3}{c}{\textbf{AutoSenCls}} \\
\cmidrule{2-10}
 & \textbf{Max $R^2$} & \textbf{Avg $R^2$} & \textbf{Ideas (i/s/t)} & \textbf{Max Acc} & \textbf{Avg Acc} & \textbf{Ideas (i/s/t)}  & \textbf{Max Acc} & \textbf{Avg Acc} & \textbf{Ideas (i/s/t)} \\
\midrule
Baseline & 27.6 & 27.6 & - & 81.2 & 81.2 & - & 91.0 & 91.0 & -\\
InternAgent (w/o AE) & 34.7 & 33.0 & 2/5/10 & 81.6 & 81.5 & 2/5/10  & 92.4 & 91.9 & 6/8/10 \\
InternAgent & \textbf{35.4} & \textbf{33.5} & \textbf{4/6/10} & \textbf{83.3} & \textbf{82.2} & \textbf{5/7/10}  &  \textbf{93.5}  & 
 \textbf{92.5}  &   \textbf{9/9/10}  \\
\bottomrule
\end{tabular}%
}
\end{table}

\subsection{Insightful Analyses}
\label{sec:insightful_analyses}

\begin{figure}[t]
    \centering
    \includegraphics[width=0.95\linewidth]{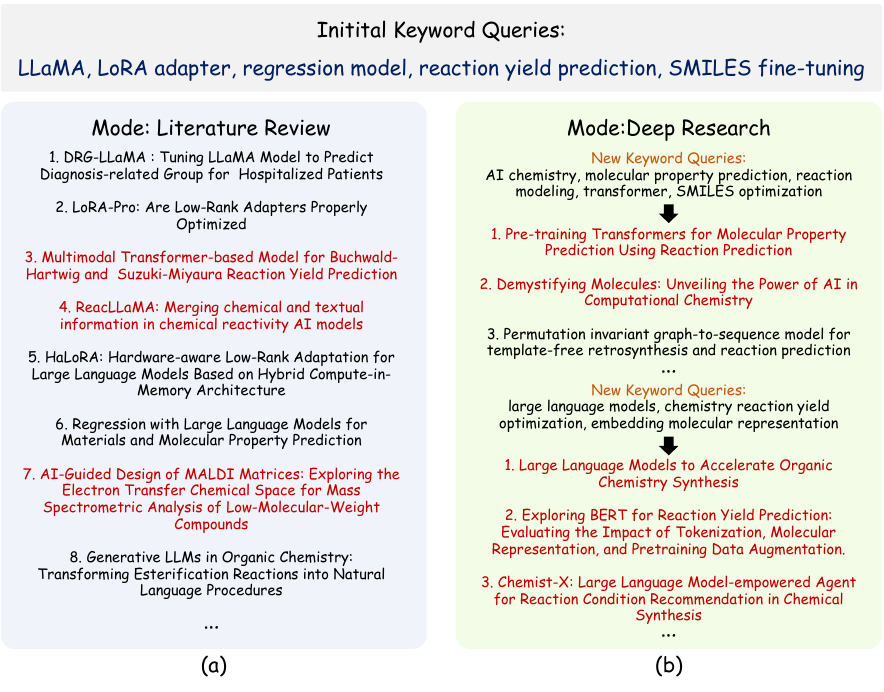}
    \vspace{-6pt}
    \caption{Analysis of two modes on survey agent.}
    \label{fig:analysis_survey_agent}
\end{figure}

\begin{table}[t]
\centering
\caption{Comparison with AI-Scientist-V2 and AI-Researcher on AutoRYP and Auto2DCls task. Total cost means the cost of the whole session. For each task, we conduct 10 experiments. AI-Scientist-V2 and AI-Researcher demonstrate relatively weak baseline improvement capabilities, with AI-Scientist-V2 in particular struggling to write code that runs correctly. The primary reason lies in the fact that AI-Scientist-V2's pipeline utilizes limited task-related information (\textit{e.g.} task formulation and type, relevant papers, and commonly used code) when generating new ideas or coding. As a result, their generated ideas tend to be more divergent and difficult to implement.}
\vspace{-8pt}
\label{tab:comparison_ai_researcher}
\renewcommand{\arraystretch}{1.0}
\resizebox{\textwidth}{!}{%
\begin{tabular}{l|cc|c|cc|c}
\toprule
\multirow{2}{*}{\textbf{Method}} &  \multicolumn{3}{c|}{\textbf{AutoRYP}} & \multicolumn{3}{c}{\textbf{Auto2DCls}}\\
\cmidrule{2-7}
 & \textbf{Max $R^2$} & \textbf{Avg $R^2$} & \textbf{Total Cost} & \textbf{Max Acc} & \textbf{Avg Acc} & \textbf{Total cost}\\
\midrule
Baseline & 27.6 & 27.6 & -   & 81.2 & 81.2   & - \\
AI-Scientist-V2~\citep{yamada2025ai} & - & - & 15\$   & - & -   & 10\$ \\
AI-Researcher~\citep{AiResearcher} & 12.3 & - & 25\$ & 80.3 & - & 32\$  \\
InternAgent & \textbf{35.4} & \textbf{33.5} & \textbf{3\$} & \textbf{83.3} & \textbf{82.2} & \textbf{3\$}  \\
\bottomrule
\end{tabular}%
}
\end{table}

\textbf{Analysis on Survey Agent.} As mentioned in Sec.~\ref{sec:survey_agent}, survey agent mainly have two modes (\textit{i.e.}, the literature review mode and the deep research mode). As shown in Fig.~\ref{fig:analysis_survey_agent} (a), under the literature review mode, the survey agent can search for domain-related papers and automatically select the most relevant literature to read and extract task-related information. For example, the agent can identify works such as "Multimodal Transformer-based Model for Buchwald-
Hartwig and Suzuki-Miyaura Reaction Yield Prediction" or "ReacLLaMA: Merging chemical and textual
information in chemical reactivity AI models" to quickly gather foundational studies in the field. Such a process is essential for idea generation process since the used agent may not have relevant domain knowledge, especially in emerging fields. Besides, under deep research mode, the survey agent needs to search for literature related to specific technical terms used in generated ideas. As shown in Fig.~\ref{fig:analysis_survey_agent} (b), the agent updates its queries based on generated technical terms and retrieves papers like "Large Language Models to Accelerate Organic Chemistry Synthesis" which are closely aligned with these refined research directions. This process is highly similar to human researchers, they initially perform a comprehensive review of the relevant field to build foundational knowledge, and then search for articles focused on specific techniques to further refine the research direction.



\textbf{Analysis on Idea Innovation Agent.} Idea innovation agent can first generate ideas and then evolve the generated ideas in an iterative manner. We take the idea evolution tree as an example to show the iterative process of polishing ideas. As shown in Fig.~\ref{fig:idea_evol}, the root node (\textit{i.e.}, Init Idea 0) denotes an initially generated idea and the child nodes are evolved from the parent node. As ideas continue to evolve, more external knowledge sourced from the survey agent is incorporated into ideas, which enriches the content and enhances the practicality of the ideas. For example, starting with a basic idea such as "adding a graph-derived reaction descriptor as a precondition for attention scores in the transformer architecture", the agent refines and evolves the ideas in an iterative manner. At each step, the evolved idea shows improvements over its predecessor in terms of technical sophistication, novelty, or practical applicability. As illustrated in Fig.~\ref{fig:idea_evol}, the process can involve incorporating more specific chemical descriptors, introducing cross-modality attention mechanisms, or leveraging hierarchical architectures, with each evolution step guided by additional insights from literature or domain knowledge, thus ensuring continuous advancement of the ideas.

\textbf{Analysis on Idea-to-Methodology Phase.} The correspondence between an idea and its final code implementation plays a crucial role in assessing the effectiveness of the idea since the idea can be verified once the experiments have been conducted. The goal of the idea-to-methodology process is to generate detailed methodologies so that code can be written based on these comprehensive method descriptions (\textit{e.g.}, method-level descriptions in research papers). As illustrated in Fig.~\ref{fig:showcase_AutoRYP}, our idea-to-method approach enables the generation of fine-grained methodologies, which facilitates accurate and faithful code implementation.

\textbf{Analysis on Evolutionary Experimental Planning and Execution.} To verify the effectiveness of the adaptive evolution (AE), we conduct ablation studies on three tasks, ranging from AI tasks to scientific tasks including AutoRYP, Auto2DCls, and AutoSenCls. As shown in Tab.~\ref{tab:ablation_study_combined}, the performance can be further improved on multiple tasks with adaptive evolution. For example, on the image classification task, both the max accuracy and the mean accuracy can be improved by 1.7\% and 0.7\%, compared to the setting without AE. This is because our coder agent can automatically analyze the previous results and baseline results and further re-plan the following experiments. Besides, the successful execution rate and the percentage of performance gains will also improve (\textit{e.g.}, on AutoRYP, the percentage of performance gains is 40\% compared to 20\% without AE). This is due to with AE, the coder will implement the idea step by step and analyze the experimental phenomena after each stage of the experiments.

\textbf{Improving Baseline in Multi-Dimension.} InternAgent not only can improve the performance on different tasks, but it also enhance the quality of ideas in other dimensions. For example, as shown in Tab.~\ref{tab:multi_dimension}, on few-shot yield prediction task. We find that the results of baseline methods are unstable, for the results of multiple repeated experiments tend to exhibit large variance (\textit{e.g.}, 24.2$\pm$4.2 when train-set=60). In contrast, the methods proposed by InternAgent can improve both the performance and the stability of the results. For example, when the train-set=60, ADAGT proposed by InternAgent achieves 34.8 average $R^2$ across 5 repeated experiments compared to 24.2 achieved by the baseline method. Besides, the variance of the results achieved by ADAGT (\textit{i.e.}, $\pm1.1$) is much lower than baseline methods (\textit{i.e.}, $\pm4.2$). This phenomenon further shows the quality of ideas and code implementation of InternAgent.

\textbf{Comparison with AI-Researcher.} We evaluate the performance and cost of InternAgent and AI-Researcher ~\citep{AiResearcher} on AutoRYP and Auto2DCls research tasks. To ensure a fair comparison, we supplied AI-Researcher with the same code templates that InternAgent uses. Both systems employ GPT-4o-2024-08-06 for idea generation and Claude-3-7-Sonnet-20250219 for code generation.  As shown in Tab.~\ref{tab:comparison_ai_researcher}, InternAgent outperforms both the baseline methods and AI-Researcher across both tasks, whereas AI-Researcher is unable to improve the provided baselines. One important reason for InternAgent's outstanding performance is its ability to generate novel ideas through extensive search and reflection.
Moreover, InternAgent has a complete experimental planning and adaptive evolution mechanism, thereby enabling it to achieve better performance.
In contrast, the idea generated by AI-Researcher is more dependent on user-provided reference papers, limiting its novelty.  Moreover, AI-Researcher often ignore the prior information of the existing codebases, which further hinders its performance. In terms of cost, InternAgent is significantly more economical than AI-Researcher. For an instance, the economic cost required for InternAgent is approximately one-sixth that of AI-Researcher on the AutoRYP task. This lower cost  allows InternAgent to conduct broader scientific experiments, thereby accelerating the exploration and validation of innovative research ideas.

%% file: sec/5_qualitative.tex
\vspace{-2pt}
\section{Case Studies}
\label{sec:case_study}
\vspace{-4pt}

\subsection{Qualitative Results}
In this section, we present the results from various tasks implemented using InternAgent.

\subsubsection{Visual Examples of Various Tasks}

\begin{figure}[t]
    \centering
    \includegraphics[width=\linewidth]{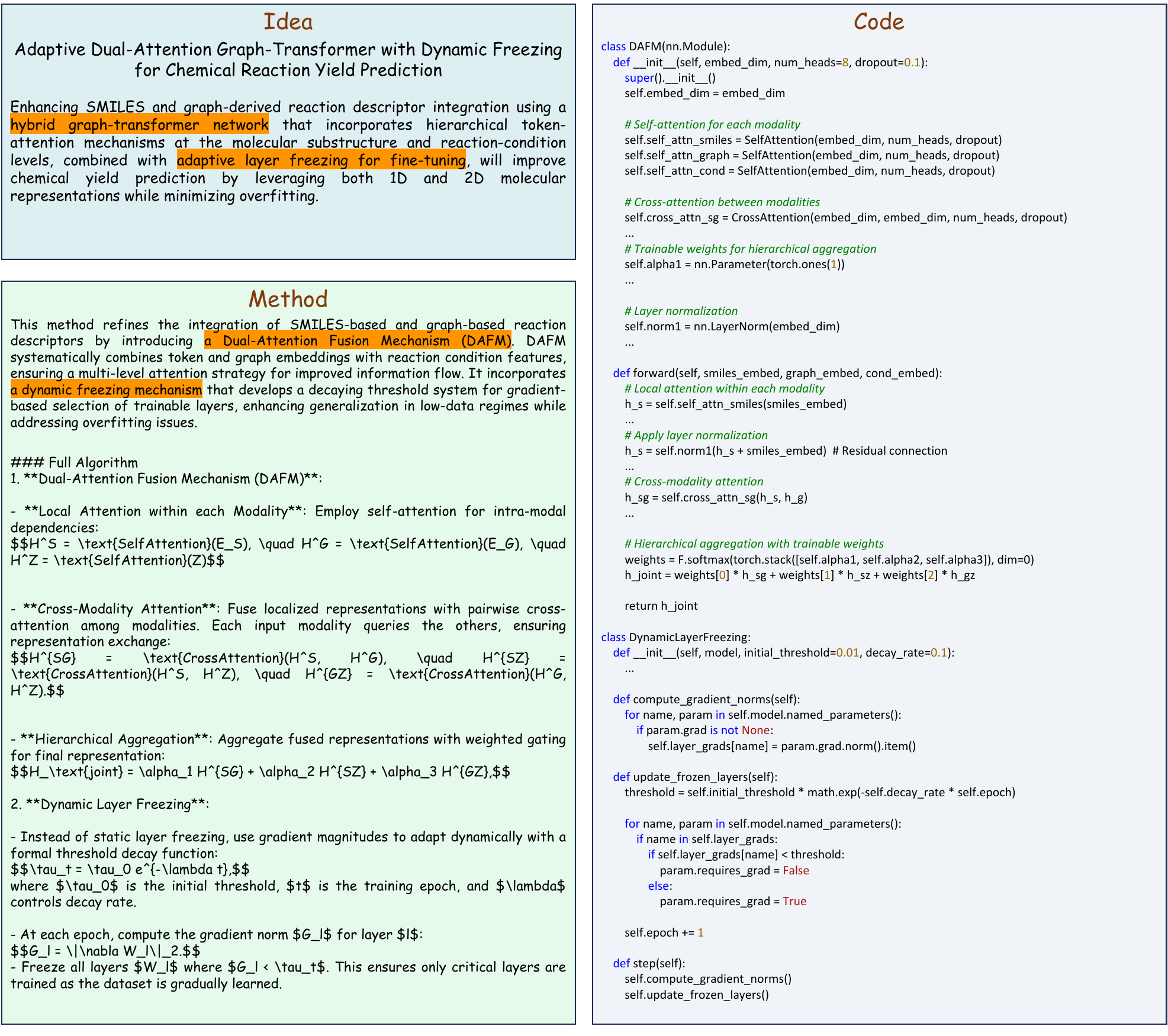}
    \vspace{-10pt}
    \caption{Visual Examples of AutoRYP Task.}
    \label{fig:showcase_AutoRYP}
\end{figure}

\begin{figure}[t]
    \centering
    \includegraphics[width=\linewidth]{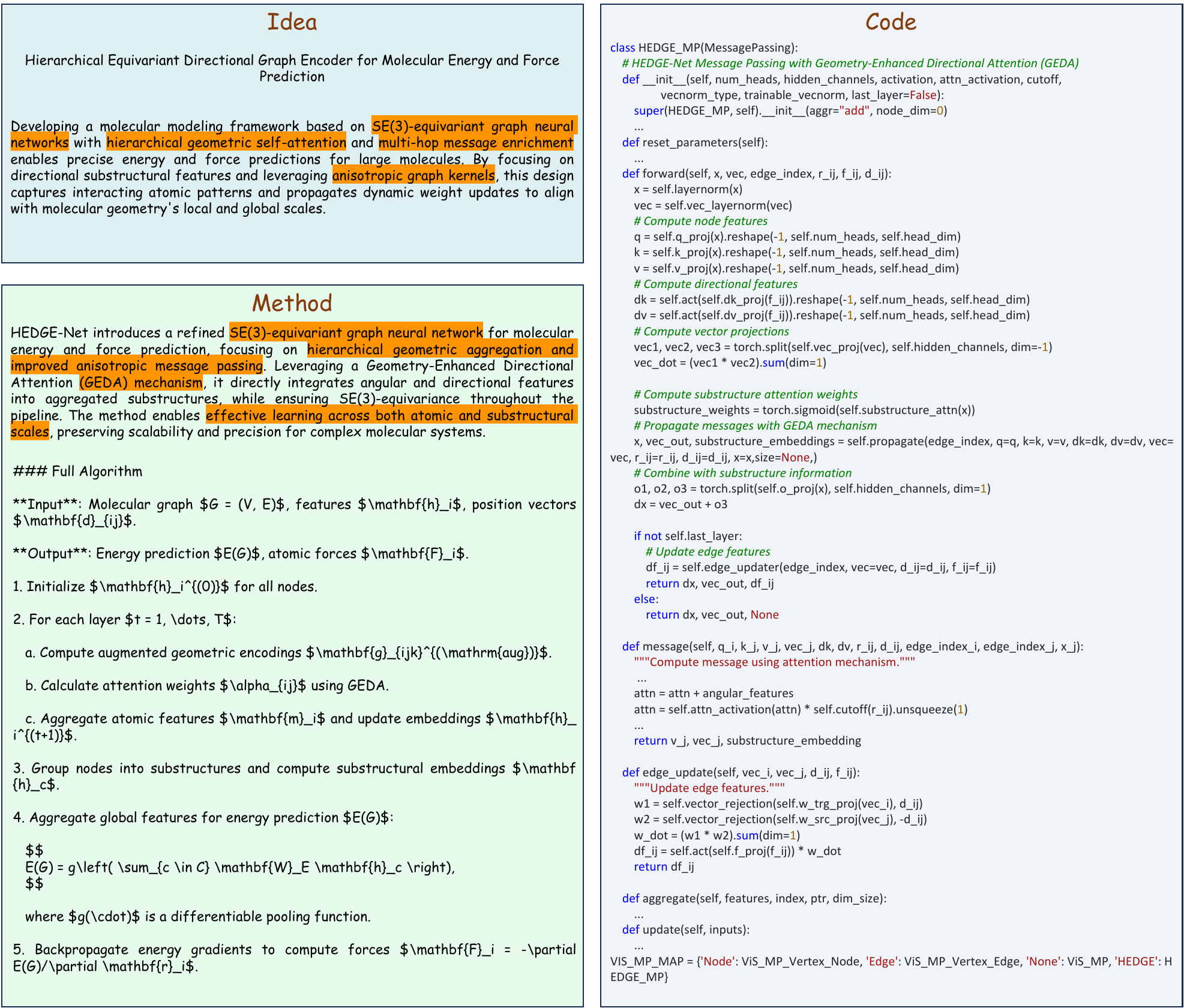}
    \vspace{-10pt}
    \caption{Visual Examples of AutoMD Task.}
    \label{fig:showcase_AutoMD}
\end{figure}

We present showcases for three distinct tasks: AutoRYP, AutoMD, and AutoPower, to highlight the innovative methodologies discovered and their applications. These showcases are illustrated in Figs. ~\ref{fig:showcase_AutoRYP}, ~\ref{fig:showcase_AutoMD}, and ~\ref{fig:showcase_AutoPower}, respectively. Each task demonstrates a unique approach to solving complex problems, showcasing the potential impact of InternAgent across different scientific domains.

In the AutoRYP task illustrated in Fig.~\ref{fig:showcase_AutoRYP}, InternAgent autonomously discovered an innovative approach called "Adaptive Dual-Attention Graph-Transformer with Dynamic Freezing" for predicting chemical reaction yields. This method effectively integrates SMILES and graph-derived descriptors using a hybrid graph-transformer network, incorporating hierarchical attention mechanisms to enhance accuracy while minimizing overfitting. The approach features a Dual-Attention Fusion Mechanism (DAFM) that systematically combines token and graph embeddings with reaction conditions, ensuring effective information flow across different representations. Furthermore, a dynamic layer freezing mechanism, based on gradient magnitudes, optimizes which layers are trained, thereby enhancing generalization in low-data scenarios. Implemented with self-attention and cross-modality attention modules, this system not only combines 1D and 2D molecular representations effectively but also improves prediction accuracy and model adaptability. Consequently, this showcase underscores the method's potential for advancing research in complex chemical tasks using deep learning.

In the AutoMD task illustrated in Fig.~\ref{fig:showcase_AutoMD}, a novel framework called "Hierarchical Equivariant Directional Graph Encoder" (HEDGE-Net) has been autonomously discovered for predicting molecular energy and forces. This approach utilizes SE(3)-equivariant graph neural networks with hierarchical geometric self-attention and multi-hop message enrichment. By integrating angular and directional features into aggregated substructures, the method captures interacting atomic patterns and propagates dynamic weight updates, aligning with both local and global molecular geometries. The core of this method, the Geometry-Enhanced Directional Attention (GEDA) mechanism, ensures SE(3)-equivariance, enabling precise predictions for complex molecular systems at both atomic and substructural scales. Implemented with advanced message passing techniques, HEDGE-Net effectively combines directional and substructural information, enhancing scalability and precision in molecular modeling. This showcases the method's potential to advance research in complex molecular tasks using deep learning techniques.

\subsubsection{Visual Examples of Experimental Planning and Adaptive  Evolution}
\label{sec:case_exp_evo}

\begin{figure}[t]
    \centering
    \includegraphics[width=0.98\linewidth]{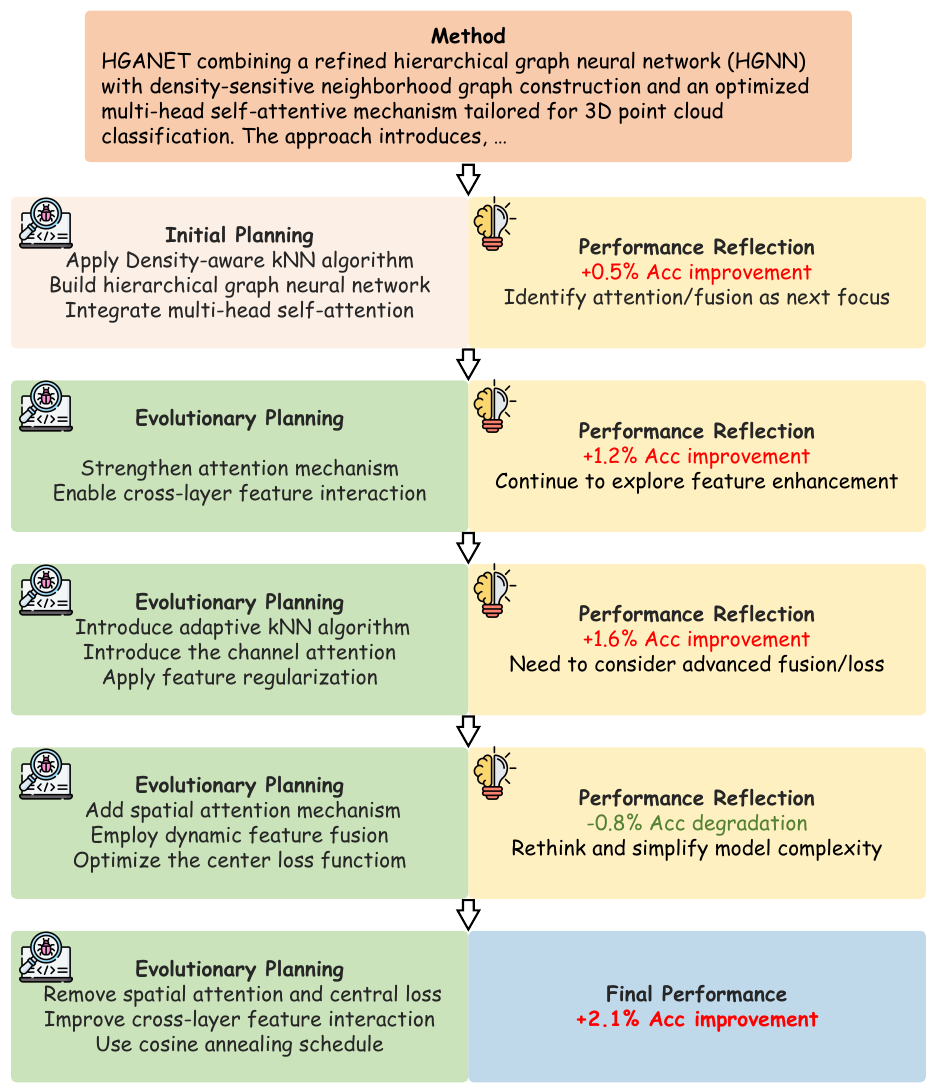}
    \vspace{-8pt}
    \caption{Visual Examples of Experimental Planning and Adaptive Evolution on Auto3DCls task.}
    \label{fig:run5_showcase_3dcls}
\end{figure}

To further illustrate the practical utility of our experimental planning and adaptive evolution framework as described in Sec.~\ref{exp_evo}, we present some concrete examples of its application in the development and optimization for 3D point cloud classification and transcription prediction for perturbation response. Fig.~\ref{fig:run5_showcase_3dcls} and Fig.~\ref{fig:run5_showcase_tppr} visually summarize the stepwise experimental planning and adaptive evolution process that guided the implementation and refinement of our method.

\subsection{Human Evaluation}

Table~\ref{tab:idea_human_eval} compares the novelty of ideas generated by our InternAgent and AI-Scientist-V2~\citep{yamada2025ai}, across various research tasks. Each task involves generating 20 ideas, which are evaluated by five qualified reviewers. The assessments focus on four criteria: soundness, contribution, overall rating, and confidence. For each research task, the average scores of the 20 ideas are reported.

In the Reaction Yield Prediction task, InternAgent outperforms AI-Scientist-V2 in all aspects, especially in overall rating and soundness. Similarly, for 2D Semantic Segmentation, InternAgent shows better idea generation ability, particularly in soundness and overall rating. In 2D Image Classification and Point Cloud Autonomous Driving, InternAgent scores higher across all criteria, indicating a consistent advantage over AI-Scientist-V2 in generating more effective and novel ideas.

\begin{table}[t]
\centering
\renewcommand{\arraystretch}{1.2}
\caption{From the perspectives of soundness, contribution, and overall, we compare the novelty of ideas generated by InternAgent and AI-Scientist-V2~\citep{yamada2025ai}. For each research task, we generate 20 ideas. Each idea is scored by 5 qualified reviewers, and the final score for each task is reported as the average score of all 20 ideas. The detailed scores for each idea can be found in the Appendix~\ref{app:criteria}.}
\vspace{-6pt}
\label{tab:idea_human_eval}
\resizebox{\textwidth}{!}{
\begin{tabular}{l|l|cccc}
\toprule
\textbf{Research Task} & \textbf{Idea-gen Method} & \textbf{Soundness} & \textbf{Contribution} & \textbf{Overall} & \textbf{Confidence}  \\
\midrule
Reaction Yield Prediction & AI-Scientist-V2 & 1.42 & 1.45 & 3.50 & 3.50 \\
Reaction Yield Prediction & InternAgent & \textbf{3.09} & \textbf{2.66} & \textbf{4.35}  & \textbf{4.00} \\ \midrule
2D Semantic Segmentation & AI-Scientist-V2 & 1.84  & 2.07  & 2.95  &  \textbf{3.64}  \\
2D Semantic Segmentation & InternAgent &  \textbf{2.41}   &  \textbf{2.35}  &  \textbf{4.05}  &  3.48   \\  \midrule
2D Image Classification & AI-Scientist-V2 & 2.78   &  2.82   &  4.40   & \textbf{3.87} \\
2D Image Classification & InternAgent & \textbf{3.15}   &  \textbf{3.10}  & \textbf{5.85}   & 3.32 \\  \midrule
Point Cloud Autonomous Driving & AI-Scientist-V2 & 2.15 & 2.47 & 3.10 & 3.94 \\
Point Cloud Autonomous Driving & InternAgent & \textbf{2.75} & \textbf{2.95} & \textbf{5.10} & \textbf{4.10} \\ 

\bottomrule
\end{tabular}
}
\end{table}

%% file: sec/6_related_works.tex
\section{Related Works}

Recent advances in Large Language Models (LLMs) and agent-based systems have demonstrated significant potential in the field of Autonomous Scientific Research (ASR), enabling progress from creative idea generation to end-to-end research automation. Some studies~\citep{li2024chain, wang2023scimon, zhou2024hypothesis} have shown that LLMs are capable of generating novel research ideas, which has sparked widespread discussion in the academic community. For example, \citet{li2024chain} introduce a method that derives research ideas through the analysis of interconnected scholarly works. Beyond idea generation, several studies have examined the use of LLMs for hypothesis formulation~\citep{qi2023large, qi2024large}, such as extracting hypotheses from large-scale web data~\citep{yang2023large}, multi-agent framework using LLMs to enhance collaborative hypothesis generation in biomedicine~\citep{qi2023large, qi2024large}, and scientific literature~\citep{wang2023scimon, zhou2024hypothesis}. However, most of these efforts remain at the stage of idea or hypothesis generation, lacking systematic empirical validation of their practical effectiveness.

In terms of end-to-end research automation, ~\citet{lu2024ai} introduced the AI Scientist framework, which was among the first to achieve a fully automated pipeline in the machine learning domain, covering problem definition, experimental execution, and result reporting. The subsequent AI Scientist-V2~\citep{yamada2025ai} further enhanced the framework by incorporating agent tree search, vision-language model feedback, and parallelized experiment execution, leading to the first workshop paper fully generated and peer-reviewed by AI. Similarly, systems such as AI-Researcher~\citep{AiResearcher} and Dolphin~\citep{yuan2025dolphin} have proposed closed-loop, LLM-driven frameworks that automate the entire research process on a range of simple tasks.

Human-AI collaboration is gaining traction in ASR. Systems like Agent Laboratory~\citep{Agentlaboratory} integrate human feedback into multi-stage LLM agent workflows, automating literature review, experiment execution, and report writing, while allowing user input at each step to enhance research quality. AgentRxiv~\citep{Agentrxiv} addresses the collaborative nature of scientific discovery by enabling LLM agent laboratories to communicate and build upon each other's work via a shared preprint server, thus facilitating knowledge sharing and collective innovation. Experimental results demonstrate that agent laboratories utilizing AgentRxiv for collaboration achieve greater performance improvements compared to isolated settings. Similarly, AI Co-Scientist~\citep{gottweis2025towards}, based on Gemini 2.0, employs a multi-agent system with a "generate-debate-evolve" strategy for hypothesis generation, and has demonstrated effectiveness in biomedical domains such as drug repurposing, novel target identification, and interpretation of bacterial evolution, with several hypotheses validated through experiments.

Despite these advances, most current systems are still evaluated primarily on relatively simple tasks or within narrow scientific domains. However, when applied to more complex, system-level scientific challenges, these approaches often face significant limitations. Key challenges include generating truly novel and scientifically sound research ideas, establishing robust closed-loop feedback between experiments and idea generation, and developing systematic evaluation standards to rigorously assess the effectiveness and real-world value of autonomous research systems.

%% file: sec/software.tex
\section{InternAgent Software Development and Application Interface}

\begin{figure}[t]
    \centering
    \includegraphics[width=\linewidth]{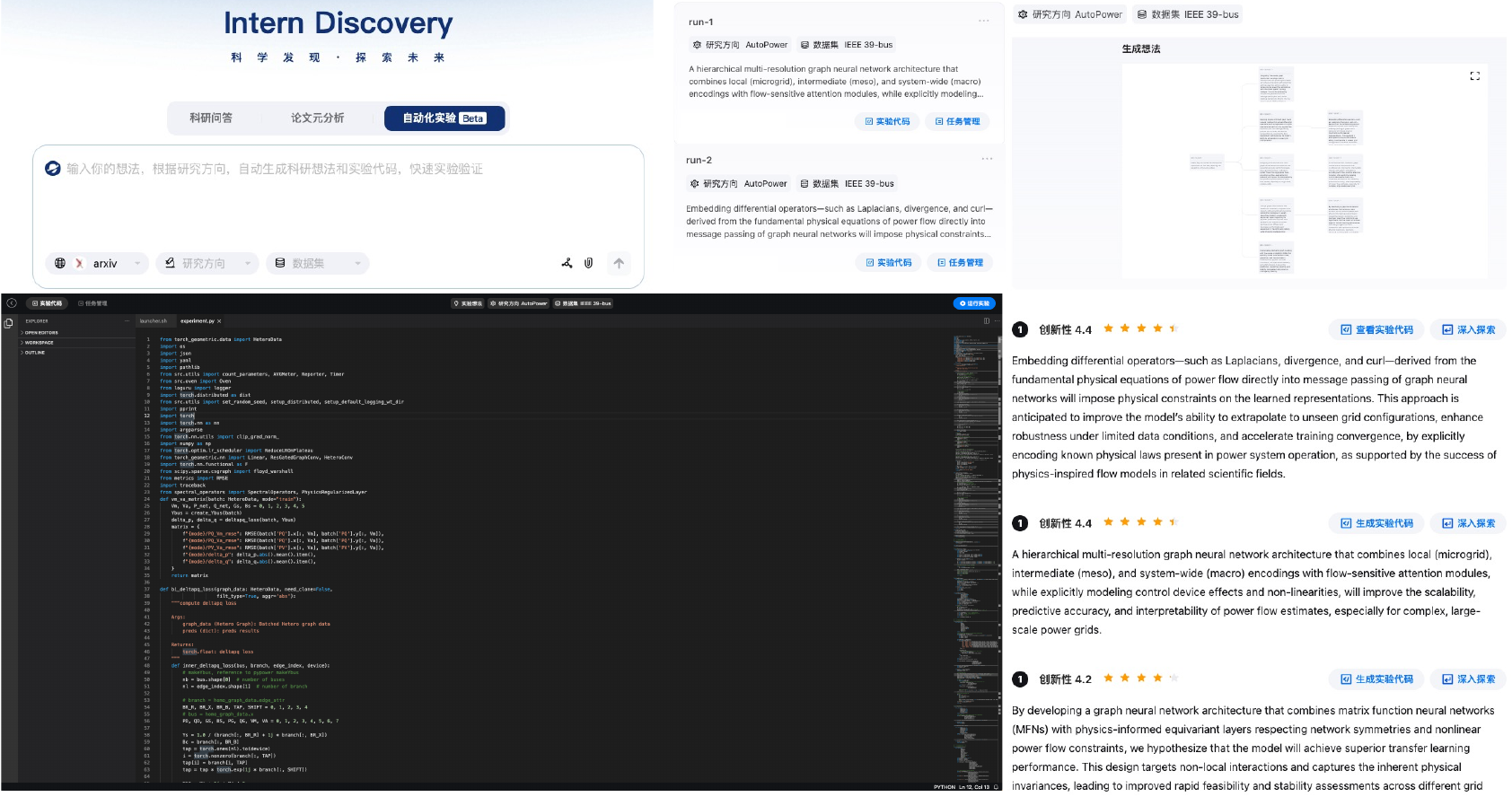}
    \vspace{-4pt}
    \caption{InternAgent software platform includes features such as the user entry interface, task selection interface, idea-tree visualization and human-computer interaction interface, code generation, and auto-debug interface. In the near future, we plan to support additional functionalities, including custom dataset uploads and academic idea thinking modes.}
    \label{fig:showcase}
\end{figure}

\noindent \textbf{Software Development.} Fig.~\ref{fig:showcase} shows the front-end interface of the current InternAgent software platform. Overall, InternAgent software platform employs a frontend-backend separation design pattern, building a highly scalable distributed service platform. 

The frontend layer is developed based on the React framework, featuring an advanced visual interaction system. Key innovations include an infinite canvas rendering engine supporting multi-node topology, a collaborative mind mapping component driven by state synchronization, a code editor supporting multiple formats, and a real-time training metrics visualization dashboard.

The backend leverages a cloud-native technology stack, utilizing a dynamic container orchestration engine for elastic resource scheduling, a distributed asynchronous task queue for high concurrency support, and a cross-cloud storage gateway for data synchronization across heterogeneous cloud environments. Additionally, a microservice governance system is established using a Service Mesh architecture.

The entire system is delivered through containerization, with Kubernetes cluster management enabling self-healing and intelligent scaling, ensuring business continuity while significantly improving resource utilization efficiency.

\noindent \textbf{InternAgent Trial Application: Accelerating Scientific Research.} InternAgent is the first end-to-end Autonomous Scientific Research (ASR) multi-agent framework. The release of this innovative achievement enables Agent to autonomously analyze problems, review literature, conduct deep research, reflect, and carry out experiments, much like human scientists. It can automatically complete the entire process from hypothesis generation to experimental validation, starting from an initial research idea. This marks the official arrival of a new era in AI-driven scientific discovery.

Due to InternAgent's reliance on massive API key calls and GPU compute clusters for executing experiments, there are associated costs. As a result, we have recently launched a whitelist application page. On this page, applicants are required to provide information about the scenarios in which they will use InternAgent, their research tasks, and the institution they belong to, so that we can select the initial batch of seed users in InternAgent. The link to the application experience page is as follows: \texttt{\url{https://discovery.intern-ai.org.cn/}}.